\documentclass[10pt,twocolumn,letterpaper]{article}

\usepackage{wacv}

\usepackage[accsupp]{axessibility}

\usepackage{times}
\usepackage{epsfig}
\usepackage{graphicx}
\usepackage{amsmath}
\usepackage{amssymb}

\usepackage{tablefootnote}

\def\wacvPaperID{1206} 

\wacvalgorithmstrack   

\wacvfinalcopy 

\ifwacvfinal
\def\assignedStartPage{1}
\fi

\ifwacvfinal
\usepackage[breaklinks=true,bookmarks=false]{hyperref}
\else
\usepackage[pagebackref=true,breaklinks=true,colorlinks,bookmarks=false]{hyperref}
\fi

\ifwacvfinal
\else
\fi

\usepackage{blindtext}
\usepackage{booktabs}
\usepackage{enumitem}
\usepackage{multirow}
\usepackage{wrapfig}
\usepackage{stfloats}
\usepackage{xcolor}
\usepackage[sort&compress,square,comma,numbers]{natbib}
\usepackage{color, colortbl}
\definecolor{Gray}{gray}{0.95}
\definecolor{LightGrey}{rgb}{0.88,0.88,0.88}
\definecolor{LightGreen}{rgb}{0.835,0.91,0.835}
\definecolor{LightYellow}{rgb}{1,1,0.4}
\definecolor{LightOrange}{rgb}{1,0.925,0.6}

\usepackage{wrapfig}
\usepackage{subfigure}
\usepackage{graphicx}

\begin{document}

\title{Bootstrapping the Relationship Between Images and Their Clean and Noisy Labels}

\author{Brandon Smart\\
Australian Institute for Machine Learning\\
University of Adelaide, Australia\\
{\tt\small a1743623@adelaide.edu.au}
\and
Gustavo Carneiro\\
Centre for Vision, Speech and Signal Processing\\
University of Surrey, United Kingdom\\
{\tt\small g.carneiro@surrey.ac.uk}
}

\maketitle

\begin{abstract}
    Many state-of-the-art noisy-label learning methods rely on learning mechanisms that estimate the samples' clean labels during training and discard their original noisy labels.
    However, this approach prevents the learning of the relationship between images, noisy labels and clean labels, which has been shown to be useful when dealing with instance-dependent label noise problems.
    Furthermore, methods that do aim to learn this relationship require cleanly annotated subsets of data, as well as distillation or multi-faceted models for training.
    In this paper, we propose a new training algorithm that relies on a simple model
    to learn the relationship between clean and noisy labels without the need for a cleanly labelled subset of data. 
    Our algorithm follows a 3-stage process, namely: 
    1) self-supervised pre-training followed by an early-stopping training of the classifier to confidently predict clean labels for a subset of the training set;
    2) use the clean set from stage (1) to bootstrap the relationship between images, noisy labels and clean labels, which we exploit for effective relabelling of the remaining training set using semi-supervised learning; and 
    3) supervised training of the classifier with all relabelled samples from stage (2). 
    By learning this relationship, we achieve state-of-the-art performance in asymmetric and instance-dependent label noise problems\footnote{Supported by Australian Research Council through grants DP180103232 and FT190100525.}. Code is available at \url{https://github.com/btsmart/bootstrapping-label-noise}.
\end{abstract}

\section{Introduction}
\label{sec:introduction}

Supervised deep learning has had great success generating effective classification models from sets of labelled training data \cite{krizhevsky2012imagenet,lecun2015deep}. 
Modern deep learning models require large-scale datasets to achieve state-of-the-art (SOTA) results \cite{pham2020meta,radford2021learning}.
However, real-world large-scale datasets, such as those collected from search engines or available from hospitals and clinics, tend to have a non-negligible amount of instance-dependent label noise (IDN)~\cite{li2017webvision, wang2017chestx}. 
Existing methods often attempt to address instance-independent label noise (IIN), such as symmetric or asymmetric noise~\cite{zhang2021learning, Goldberger2017TrainingDN, xiao2015learning}. Handling the IDN present in large-scale real-world datasets has become one of the main research problems in the field.

When naively trained with noisy-labelled data, deep learning models generalise poorly because they can easily overfit the incorrectly labelled samples~\cite{zhang2016understanding}. 
Many methods have been developed for handling label noise, with SOTA approaches relying on sample relabelling mechanisms. 
These strategies are based on techniques to estimate the relationship between images and clean labels, and after relabelling, the old noisy labels are discarded~\cite{li2020dividemix, zhang2021learning, song2019selfie}.
However, to model how different image features and noisy labels affect the mislabelling process in IDN, we need to estimate the relationship between images, clean labels and noisy labels~\cite{zhang2021learning, gu2021instancedependent}.
Some methods have attempted to model this relationship with noise-transition matrices and corrective layers for asymmetric noise~\cite{xia2019anchor, Goldberger2017TrainingDN, patrini2017making} or part-dependant noise in place of instance-dependant noise~\cite{xia2020part}, but they failed to achieve SOTA results.

\begin{figure}[t!]
    \begin{center}
    \includegraphics[width = 6.3cm]{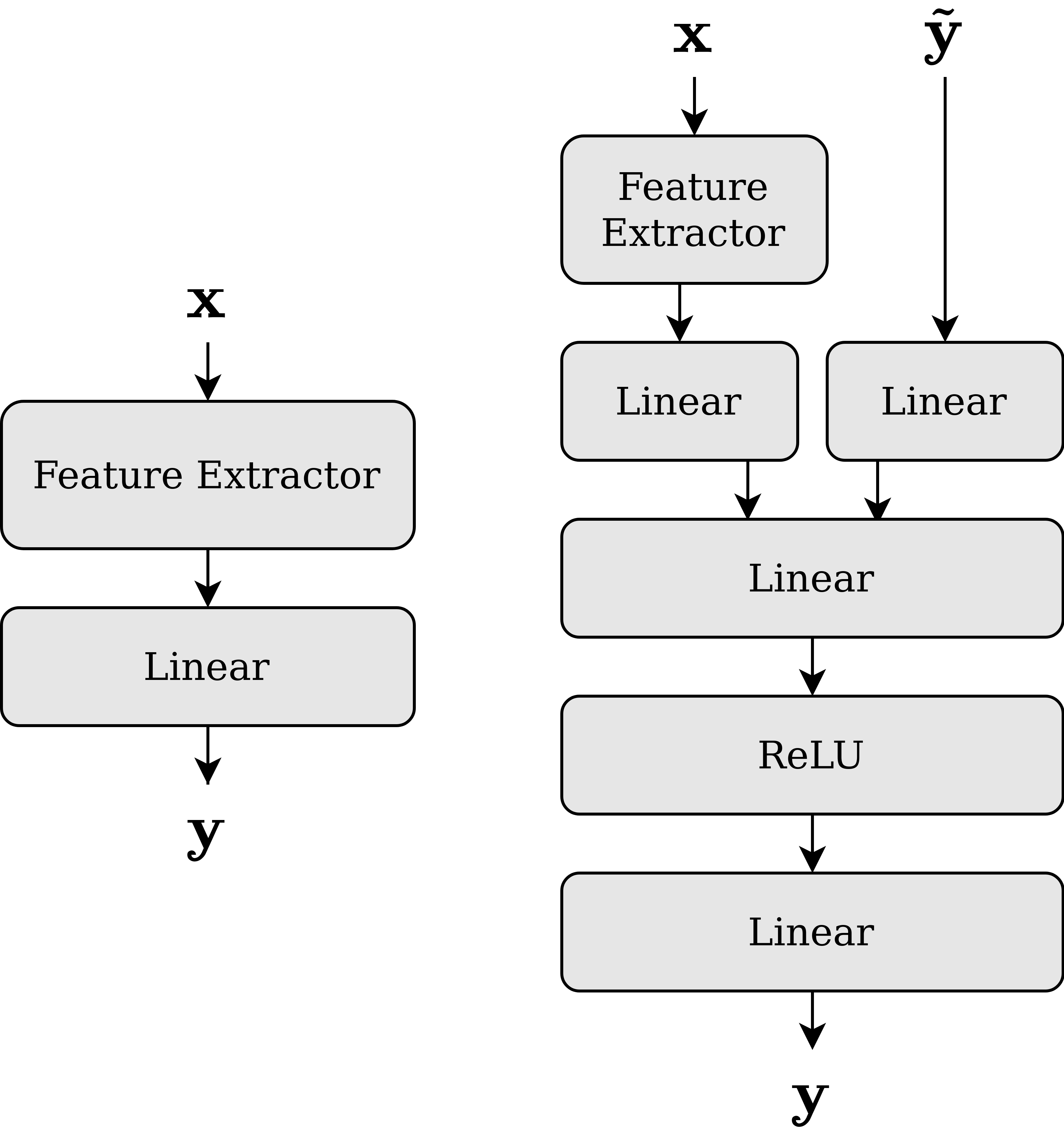}
    \end{center}
    \caption{\small On the left, we show a `normal' deep neural network model used for noisy label learning tasks. On the right, we present a `modified' model that can learn the relationship between images $\mathbf{x}$, noisy labels $\tilde{\mathbf{y}}$ and clean labels $\mathbf{y}$, similar to those used by methods that have access to a clean set of data~\cite{inoue2017multi,veit2017learning,gu2021instancedependent}.}
    \label{fig:models}
\end{figure}

\begin{figure*}[t!]
    \begin{center}
    \includegraphics[width = 17.0cm]{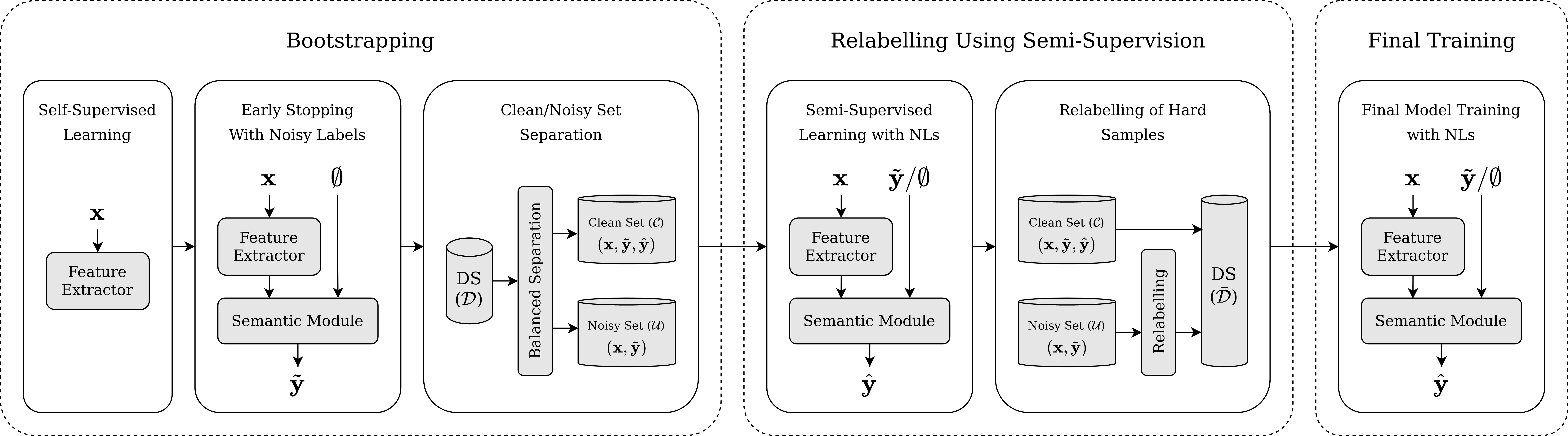}
    \end{center}
    \caption{\small Our proposed algorithm. The bootstrapping stage consists of self-supervised pre-training, followed by early-stopping classification training to identify a small set of confidently relabelled samples (i.e., clean labelled samples).
    This set is then used to learn the relationship between images, noisy labels and clean labels during semi-supervised learning, and at the end of this stage, we relabel the samples classified as noisy during bootstrapping. The last stage is the final training that trains the classifier using the clean and noisy samples identified in the SSL stage.}
    \label{fig:methodology}
\end{figure*}

Rather than the usual noisy-label learning setting, where a set of cleanly annotated samples is not available, some methods assume the existence of a subset of training data containing images, clean labels \textit{and} noisy labels~\cite{inoue2017multi,veit2017learning,gu2021instancedependent}. By training a model that predicts clean labels from both images and noisy labels (see right of Figure \ref{fig:models}), these methods are able to learn the relationship between image features, noisy labels and clean labels, allowing them to model IDN and more effectively relabel noisy samples. However, it can be expensive, difficult and time-consuming to obtain a clean subset of data with noisy labels \textit{and} clean labels that is representative of the instance-dependant noise in the dataset. 
Furthermore, these methods require distillation to a more standard model (such as the one on the left of Figure~\ref{fig:models}) for evaluation on samples without labels.

In this paper, we introduce a new algorithm to learn the relationship between images and their clean and noisy labels without using any clean-label set.
Our algorithm follows a 3-stage process (see Fig.~\ref{fig:methodology}): 
1) \textit{Bootstrapping:} self-supervised pre-training followed by an early-stopping training~\cite{zhang2016understanding} of the classifier that receives images and `null' labels as input and predicts the noisy labels as output -- this stage forms a subset of predicted clean labels for the second stage of training;
2) \textit{Semi-supervised Learning:} use this predicted clean subset to learn the relationship between images, noisy labels and clean labels, which we exploit for an effective, explicit relabelling of the remaining training set;
and 3) \textit{Final training:} supervised training of the classifier using the relabelled samples. The main contributions of this paper are:
\begin{itemize}
    \item An effective three-stage training algorithm designed to address instance-dependent label noise by learning the relationship between images and their clean and noisy labels -- using a noise-transition sample balancing scheme, explicitly relabelling training samples and without requiring a cleanly annotated training set;
    \item A method that reaches SOTA asymmetric and instance dependent label noise results using a simple single-model architecture, unlike DivideMix \cite{li2020dividemix} (and its derivatives such as \cite{sachdeva2021scanmix, zheltonozhskii2022contrast, Nishi_2021_CVPR, cordeiro2021propmix, kim2021fine}) that require a more complex 2-model architecture.
    \item A `label dropping' strategy that removes the need for distillation to a standard model and allows predictions to be made on samples with \textit{and} without noisy labels;
\end{itemize}

\section{Related Work}

\subsection{Noisy Label Learning Based on Semi-supervised Learning (SSL)}
\label{sec:related_work_SSL}

Many SOTA methods in noisy label learning use SSL techniques to perform label correction and consistency regularization. DivideMix~\cite{li2020dividemix} and variants~\cite{sachdeva2021scanmix, cordeiro2021propmix, zheltonozhskii2022contrast} perform sample re-labelling with co-teaching and \textit{MixMatch} data augmentation~\cite{berthelot2019mixmatch}. 
ELR+~\cite{liu2020early} and PES~\cite{bai2021understanding} similarly use MixUp \cite{zhang2018mixup} based SSL on top of a regularising loss functions. 
These techniques are effective for symmetric and asymmetric noise, but are dependent on carefully tuned hyperparameters 
and cautious integration of the sample re-labelling and SSL techniques used.

To identify the incorrectly labelled samples for forming the unlabelled set, many methods depend on loss-separation techniques, relying on the ability of deep networks to learn clean samples faster than noisy samples~\cite{arpit2017closer}, which leads to lower loss values for clean samples after a few stages of  training~\cite{han2018co, chen2019understanding, arazo2019unsupervised, li2020dividemix, xu2021faster}. FINE alternatively uses eigen-decomposition to separate samples in feature space \cite{kim2021fine}.  
However, such automatic and dynamic identification of noisy-label samples is a brittle process that tends to fail in challenging noisy-label learning scenarios, such as instance-dependent label noise, because the differences between hard clean-label and noisy-label samples can be subtle during early training stages.

\subsection{Label Transition Estimation Methods}

Many methods attempt to model the class-dependent asymmetric noise, such as with a label transition matrix~\cite{xia2019anchor}, by learning
noise-adaption layers and performing loss-correction \cite{Goldberger2017TrainingDN, patrini2017making},
or by using reconstruction error as a consistency objective~\cite{reed2014training}. However, these methods are not as  competitive as SSL approaches in Sec.~\ref{sec:related_work_SSL} because they generally do not address instance-dependent noise and have 
limited mechanisms to make use of mislabeled samples.
Methods that attempt to handle semantic noise by estimating instance-based transition matrices can in principle deal with semantic noise~\cite{Goldberger2017TrainingDN} and can provide guarantees on convergence and bounds on generalization error~\cite{zhang2021learning}, but they do not provide SOTA results in practice.

\subsection{Methods Based on Clean Validation Sets}

Alternatively, researchers have explored learning methods that require the existence of a small, additional clean set of data to learn from. For instance, many meta-learning strategies require clean validation samples to adjust the weights of each training sample~\cite{ren2018learning}, to simulate regular training with synthetic noise labels~\cite{li2019learning}, to learn an explicit weighting function~\cite{shu2019meta}, or to estimate the noise transition matrix~\cite{wang2020training}.
Other noisy label learning algorithms rely on clean sets of data for which noisy labels \textit{and} clean labels exist for samples, so that the relationship between image features, noisy labels and clean labels can be learned~\cite{veit2017learning, inoue2017multi, gu2021instancedependent}, using fully-connected neural networks to predict true labels from images and their noisy labels. 
Together, these methods show the utility of representative clean sets of data to the noisy label problem, but they rely on manual labelling, which can be expensive and time-consuming to collect, particularly in domains that require a high degree of expertise for labellers, such as medical imaging~\cite{lloyd2004observer, zhang2020distilling}.

Other methods aim to dynamically construct a pseudo-clean set out of high-confidence samples, such as by using K-Nearest Neighbours to identify related samples in the feature space~\cite{bahri2020deep, ortego2020multi}, or by using meta-learning to identify a dictionary of dynamically updating valuable training samples~\cite{zhang2021learning3}.

\subsection{Background Material}

Our method relies on many techniques previously developed in the field, such as self-supervision, SSL and data augmentation. Recently, self-supervised methods such as SimCLR \cite{chen2020simple, chen2020big} and SCAN \cite{van2020scan} have been used for pre-training, or as auxilliary objectives in noisy-label learning tasks \cite{zheltonozhskii2022contrast, sachdeva2021scanmix}, due to their ability to learn high-level features from noisy data without the risk of overfitting incorrect labels. 
In this paper, we utilize the SSL method FixMatch~\cite{sohn2020fixmatch}, which uses pseudo-labelling thresholds and consistency between strong and weak augmentations to regularize training through consistency regularization~\cite{bachman2014learning, tarvainen2017mean, Laine2017TemporalEF} and entropy minimization~\cite{grandvalet2004semi, lee2013pseudo}.
Strong data augmentation strategies, such as RandAugment~\cite{cubuk2020randaugment}, AutoAugment~\cite{cubuk2019autoaugment} and MixUp~\cite{zhang2018mixup} have been shown to be effective for regularising training, preventing overfitting and dramatically improving the tolerance of algorithms to higher noise levels~\cite{Nishi_2021_CVPR, li2020dividemix, arazo2019unsupervised}.

\section{Methodology}
\label{sec:methodology}

Our algorithm (see Fig.~\ref{fig:methodology}) is motivated by the objective to train a model that can accurately relabel samples by predicting true labels from images \textit{and} noisy labels without requiring clean-labelled data. The stages of the proposed algorithm are: 1) Bootstrapping: perform self-supervised pre-training and early-stopping training to identify a representative, clean subset of samples, 2) SSL: learn the instance-dependant noise relationship from the clean set (from stage 1) and use it to relabel the remaining noisy samples, and 3) Final Training: use the relabelled samples from stage 2 to train a final, regularized classifier.

For the methods described below, assume the availability of a training set $\mathcal{D}=\{(\mathbf{x}_i,\tilde{\mathbf{y}}_i)\}_{i=1}^{|\mathcal{D}|}$, where $\mathbf{x} \in \mathcal{X} \subset \mathbb{R}^{H \times W \times R}$ denotes an image of size $H \times W$ with $R$ colour channels, and $\tilde{\mathbf{y}} \in \mathcal{Y} \subset \{0,1\}^{|\mathcal{Y}|}$ represents the one-hot noisy label.
Our model, referred to as 'modified', receives an image and the noisy label at the input and outputs a clean label classification distribution, with $f_\theta:\mathcal{X} \times \mathcal{Y} \to \Delta_{|\mathcal{Y}|-1}$, where $\Delta_{|\mathcal{Y}|-1}$ represents the $|\mathcal{Y}|-1$ probability simplex, and $\theta \in \Theta$ denotes the model parameters. Note that we also consider a `normal' model, which is a model that takes an image input and outputs a classification, with $f_\theta:\mathcal{X} \to \Delta_{|\mathcal{Y}|-1}$.

\subsection{Bootstrapping}
\label{sec:bootstrapping}

The goal of this first stage is to train a model that accepts images and noisy labels and predicts clean labels, however at the beginning of training we only have access to images and noisy labels from $\mathcal{D}$. 
Following~\cite{van2020scan, zheltonozhskii2022contrast}, we start with SimCLR pre-training~\cite{chen2020simple}, which allows us to learn an initial feature representation from $\mathcal{D}$ without the risk of overfitting.

Next, we take the pre-trained model above to learn a classifier with early-stopping and small learning rate with
\begin{equation}
\scalebox{0.85}{$
    \theta^* = \arg\min_{\theta}\frac{1}{|\mathcal{D}|}\sum_{(\mathbf{x}_i,\tilde{\mathbf{y}}_i) \in \mathcal{D}} \mathbb{E}_{a(.) \sim \mathcal{A}_S}\left [\ell_{CE}(\tilde{\mathbf{y}}_i,f_{\theta}(a(\mathbf{x}_i),\mathbf{0}_{|\mathcal{Y}|}))\right ],
    $}
    \label{eq:train_strong_aug}
\end{equation}
where $a(.)$ is a strong data augmentation sampled from the set of strong data augmentation functions $\mathcal{A}_S$,  $\ell_{CE}(.)$ denotes the cross-entropy loss function, and $\mathbf{0}_{|\mathcal{Y}|}$ is a `null' label vector with $|\mathcal{Y}|$ zeros.

Then, we use our trained model to generate a prediction distribution for all training samples using test-time weak augmentation, as follows:
\begin{equation}
    \hat{\mathbf{y}}_i = \mathbb{E}_{a(.) \sim \mathcal{A}_W}\left [ f_{\theta^{*}}(a(\mathbf{x}_i),\mathbf{0}_{|\mathcal{Y}|}) \right ],
    \label{eq:evaluate_weak_aug}
\end{equation}
where $a(.)$ is a weak augmentation sampled from the set of weak data augmentation functions $\mathcal{A}_W$. We also have dropout enabled during this evaluation process. By using dropout and multiple weak augmentations to evaluate samples, we penalise samples with highly confident but inconsistent predictions~\cite{zhang2021learning}. 
The confidence prediction for $\mathbf{x}_i$ is given by $\max_{c \in \mathcal{Y}} \hat{\mathbf{y}}_i(c)$. 

We then split the dataset into a confident clean set and a noisy set. However, if we naively select the most confident samples, we will disproportionally select samples from easy classes, and samples whose original noisy labels were correct.
We want the clean set to contain representative samples from all classes and noise transitions in the dataset, as the upcoming SSL process can only learn noise transitions which are present in the clean set. 
To achieve this, we propose noise-transition sample balancing that
first estimates the dataset's noise transition matrix $\mathbf{T}$ by using the noisy labels and predicted labels for the 90\% of most confident predictions per class, where $\mathbf{T}_{ij}$ represents the estimated proportion of samples in the dataset with the noisy label $i$ and clean label $j$. 
We then select the $K \times |\mathcal{Y}| \times \mathbf{T}_{ij}$ most confident samples from each noise transition, as well as any other samples $\mathbf{x}_i$ where $\max_{c \in \mathcal{Y}} \hat{\mathbf{y}}_i(c) > \tau$, where $K$ is a hyperparameter controlling the minimum fraction of samples from each subset we want to select, and $\tau$ is a hyperparameter controlling how confident a prediction needs to be before it is guaranteed to be selected.

We note that this process does not guarantee that all instance-dependent relationships between image features and noisy label transitions are captured, but in practice this ensures a large coverage of the different noise transitions present in the dataset. This `noise-based' balancing approach can be contrasted against the usual class-based balancing typically seen in noisy-label learning methods, where samples are selected to balance the number of samples per class.
The initial clean set will contain the samples with both the noisy and estimated clean labels with $\mathcal{C} = \{(\mathbf{x}_i,\tilde{\mathbf{y}}_i,\hat{\mathbf{y}}_i) | (\mathbf{x}_i,\tilde{\mathbf{y}}_i) \in \mathcal{D}\}$, and the initial noisy set will contain the samples and noisy labels as $\mathcal{U}=\{ (\mathbf{x}_i,\tilde{\mathbf{y}}_i) |  (\mathbf{x}_i,\tilde{\mathbf{y}}_i,\hat{\mathbf{y}}_i) \notin \mathcal{C} \text{, and } (\mathbf{x}_i,\tilde{\mathbf{y}}_i) \in \mathcal{D}\}$.

\subsection{Semi-Supervised Learning (SSL) for Noisy Label Correction}

\begin{figure}[ht!]
    \begin{center}
    \includegraphics[width = 8.0cm]{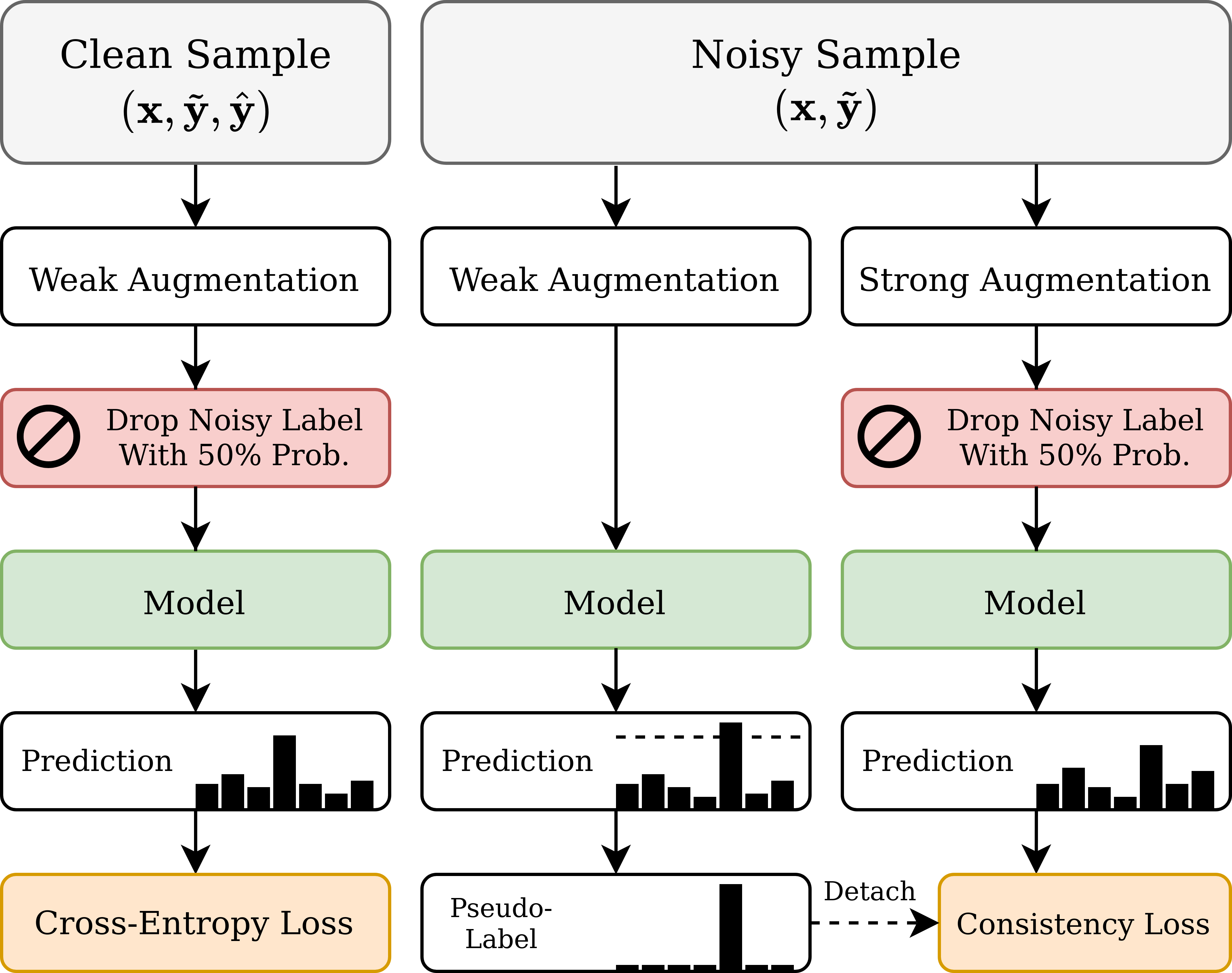}
    \end{center}
    \caption{\small The noisy-label  FixMatch algorithm.}
    \label{fig:fixmatch}
\end{figure}

The next stage of our framework takes the initial clean set $\mathcal{C}$ and the initial noisy set $\mathcal{U}$ 
to train the SSL model, where images and noisy labels (the model inputs) are present for all samples, and the true labels (the model output) are present for samples in $\mathcal{C}$. 
Our SSL algorithm is based on FixMatch (see Fig.~\ref{fig:fixmatch}), which achieves competitive performance by focusing on consistency regularization and entropy minimization~\cite{sohn2020fixmatch}. We do not reinitialize the network before semi-supervised learning, instead using the bootstrapping process as a form of warmup.

Because we are learning to predict the `true' labels of samples from images and noisy labels, our model is able to learn a joint distribution between the three, similar to works that require a clean set ~\cite{veit2017learning, inoue2017multi, gu2021instancedependent}. However, if we were to train our model to always make predictions from images \textit{and} noisy labels, our model would no longer be able to make meaningful predictions on samples without associated noisy labels, such as those in a `test' set. To remedy this, we randomly `drop' the one-hot noisy label from samples 50\% of the time, replacing it with a `null' label. By doing this, our model also learns a direct relationship between images and true labels without depending on the noisy labels, allowing us to evaluate test samples by passing them into the model alongside a `null' label. In our implementation of FixMatch, we drop the noisy label of supervised samples and strong augmentations of unsupervised samples 50\% of the time. However, we keep the noisy labels for the weakly augmented unsupervised samples as we want to always use these noisy labels to predict higher accuracy pseudo-labels, given that the loss is not backpropagated along these samples (see Fig.~\ref{fig:fixmatch}). We experiment with this decision in the Supplementary Material.
The SSL training is based on the following optimisation:
\begin{equation}
\scalebox{0.8}{$
\begin{aligned}
    \theta^{*} & = \arg\min_{\theta \in \Theta} \\
    &\frac{1}{|\mathcal{C}|} \sum_{(\mathbf{x}_i,\tilde{\mathbf{x}}_i,\hat{\mathbf{y}}_i)\in\mathcal{C}} \mathbb{E}_{a(.)\in\mathcal{A}_W}[ \ell_{CE}(\hat{\mathbf{y}}_i,f_{\theta}(a(\mathbf{x}_i),\iota_{50\%}(\tilde{\mathbf{y}}_i)))] + \\
    &\frac{1}{|\mathcal{U}|} \sum_{(\mathbf{x}_i,\tilde{\mathbf{x}}_i)\in\mathcal{U}}\mathbb{I}(\max \bar{\mathbf{y}}_i>\kappa)\mathbb{E}_{a(.)\in\mathcal{A}_S}[ \ell_{CE}(\lceil \bar{\mathbf{y}}_i \rceil,f_{\theta}(a(\mathbf{x}_i),\iota_{50\%}(\tilde{\mathbf{y}}_i))) ],
\end{aligned}
$}
\label{eq:SSL}
\end{equation}
where $\mathbb{I}(.)$ denotes an indicator function, $\bar{\mathbf{y}}_i = \mathbb{E}_{a(.) \sim \mathcal{A}_W}[f(a(\mathbf{x}_i),\tilde{\mathbf{y}}_i)]$, 
$\iota_{50\%}(\tilde{\mathbf{y}}_i)$ randomly returns $\tilde{\mathbf{y}}_i$ or $\mathbf{0}_{|\mathcal{Y}|}$, each with $50\%$ chance,
and 
$\lceil \bar{\mathbf{y}}_i \rceil$ is an operator to transform $\bar{\mathbf{y}}_i$ into a binary vector, with $1$ for the class with largest probability and $0$ otherwise.
After this SSL stage, we re-label the whole training set to form
\begin{equation}
\bar{\mathcal{D}} = \{(\mathbf{x}_i, \tilde{\mathbf{y}}_i, \bar{\mathbf{y}}_i) | (\mathbf{x}_i, \tilde{\mathbf{y}}_i)\in \mathcal{D}\},
\label{eq:relabel_D}
\end{equation}
with $\bar{\mathbf{y}}_i$ defined in~\eqref{eq:SSL}. As in the bootstrapping phase of training, we enable dropout when averaging over the predictions of multiple weak augmentations to calculate $\bar{\mathbf{y}}_i$. Note in~\eqref{eq:relabel_D} that our model uses the learned relationship between images, noisy labels and `true' labels to relabel the remaining noisy samples.

\subsection{Final Model Training}

After forming $\bar{\mathcal{D}}$, we train a final model with strong augmentations and MixUp, due to their robustness to any noise that may remain in the re-labelled training set~\cite{arazo2019unsupervised, li2020dividemix}. MixUp is applied on both images and their noisy labels together. After applying Mixup, we randomly replace the noisy label with a `null' label in 50\% of samples.

\section{Experiments}

\subsection{Data sets}

To investigate our method, we perform experiments using the CIFAR-10, CIFAR-100~\cite{krizhevsky2009learning}, Animal10N~\cite{song2019selfie}, 
and Webvision~\cite{li2017webvision} datasets.
CIFAR-10 and CIFAR-100 consist of 50,000 training and 10,000 testing images of size $32 \times 32$ pixels, with 10 and 100 classes respectively. As CIFAR-10 and CIFAR-100 do not contain label noise, we follow the literature and perform experiments with different types of controlled, synthetic label noise. The first type of noise is the Polynomial Margin Diminishing (PMD) semantic noise~\cite{zhang2020learning}, where confusing samples near decision boundaries are mislabelled at higher rates than samples far from decision boundaries. The second type is the semantic noise introduced by Lee et al.~\cite{lee2019robust}, where incorrect predictions from trained VGG~\citep{simonyan2014very}, DenseNet~\cite{huang2017densely}, and ResNet~\cite{he2016identity} models are used to generate mislabelled samples (which we refer to as `RoG' noise). We also test our system with symmetric noise rates of $\{20\%, 50\%, 80\%, 90\%\}$ and asymmetric noise using the mapping from~\citep{li2020dividemix, patrini2017making} with $40\%$ rate.

The Animal10N dataset~\cite{song2019selfie} consists of 50,000 training images and 5,000 testing images of size $64 \times 64$, consisting of five pairs of semantically similar classes. Images are collected for each label using online search engines, resulting in incorrect classifications for an estimated $6\%$ to $10\%$ of samples.
Finally, we test with Mini-Webvision, which consists of the 65,944 samples which make up the first 50 classes of the Webvision dataset that contains images collected from the internet. Images are resized to $256 \times 256$, and the corresponding 50 classes in the ILSVRC12 dataset \cite{deng2009imagenet} are also used for validation.

\subsection{Implementation}

Following contemporary work~\citep{li2020dividemix, cordeiro2021propmix, liu2020early}, we use a PreAct-ResNet-18 (PRN18) network~\citep{he2016identity} as our backbone model for CIFAR10 and CIFAR100 experiments, use a VGG19 model~\cite{simonyan2014very} as our backbone for Animal10N, and use a Inception-ResnetV2 model~\cite{szegedy2017inception} as our backbone for Webvision. For weak augmentations, we use horizontal flipping and random cropping, and for strong augmentations we used AutoAugment \cite{cubuk2019autoaugment}, followed by horizontal flipping and random cropping. For all experiments, we perform bootstrapping with strong augmentations 
and perform pseudo-labelling with with 25 weak augmentations. Final model training is done using Mixup \cite{zhang2018mixup} (with $\alpha=1$). For all stages of training, we use stochastic gradient descent, with additional information about the optimizer and training schedule hyperparameters provided in the supplementary material. For Webvision, we additionally use label smoothing (with $\epsilon=0.1$).
Following existing implementations of FixMatch (such as by \textit{TorchSSL} \cite{zhang2021flexmatch}), we use Exponential Moving Average (EMA) models to perform temporal ensembling \cite{Laine2017TemporalEF}. For fair comparison with existing noisy-label learning methods, we shorten the training schedules typically used by FixMatch implementations (we use 100,000 training iterations with $\mu=4$ rather than 1,000,000 iterations with $\mu=8$). 

We report two sets of results for our experiments in order to understand how the use of noisy labels during semi-supervised learning can improve results: 'Normal Model' and 'Modified Model'. In both cases, we utilise the training procedure outlined in Figure \ref{fig:methodology}, however:
\begin{enumerate}[label=\arabic*)]
    \item In `Normal Model', we use a standard model which only accepts image inputs. Because this type of model does not aim to learn the relationship between noisy labels and clean labels, we perform class-based balancing rather than the proposed noise-based balancing.
    \item In `Modified Model', we use 
    noise-based balancing, as well as the  model that accepts images and noisy labels, as described in Sec.~\ref{sec:methodology}.
\end{enumerate}
For all our experiments, we also report the accuracy obtained from test-time augmentation, where 25 weakly augmented versions of each testing sample are generated, and the model's predictions (with dropout enabled) are averaged to generate a final prediction. Because this extends the inference time for our model, we report these results separately in italics.

\subsection{Results}

\begin{table}[t!]
    \begin{center}
    \scalebox{0.8}{
        \begin{tabular}{lcc}
            \toprule
            {Method} & {Top-1} \\ 
            \midrule
            Cross Entropy & 79.4 \\
            SELFIE \cite{song2019selfie} & 81.8 \\
            PLC \cite{zhang2020learning} & 83.4 \\
            NCT \cite{sarfraz2021noisy} & 84.1 \\
            \midrule
            Ours (Normal Model) & 85.84 \\
            {\hskip 0.5cm} + \textit{Test-Time Aug.} & \textit{86.98} \\
            Ours (Modified Model) {\hskip 0.1cm} & \textbf{88.48} \\
            {\hskip 0.5cm} + \textit{Test-Time Aug.} & \textit{89.38} \\
            \bottomrule
        \end{tabular}
    }
    \end{center}
    \caption{Test accuracy (\%) for Animal10N. Top methods within $1\%$ in \textbf{bold} (Results other than ours are from \cite{sarfraz2021noisy}).}
    \label{tab:animal10n_2}
\end{table}

\begin{table}[t!]
    \begin{center}
    \scalebox{0.8}{
        \begin{tabular}{lcccc}
            \toprule
            \multicolumn{1}{c}{Method} & \multicolumn{2}{c}{Webvision} & \multicolumn{2}{c}{ILSVRC2012} \\
            \multicolumn{1}{c}{} & Top-1 & Top-5 & Top-1 & Top-5 \\
            \midrule
            ELR~\cite{liu2020early} & 76.26 & 91.26 & 68.71 & 87.84 \\
            ELR+~\cite{liu2020early} & 77.78 & 91.68 & 70.29 & 89.76 \\
            PropMix~\cite{cordeiro2021propmix} & 78.84 & 90.56 & / & / \\
            NGC~\cite{wu2021ngc} & 79.16 & 91.84 & 74.44 & 91.04 \\
            FaMuS~\citep{FaMUS} & 79.40 & \textbf{92.80} & \textbf{77.00} & \textbf{92.76} \\
            RRL~\cite{li2021learning} & 76.30 & 91.50 & 73.30 & 91.20 \\
            \midrule
            \textbf{Ours (Normal Model)} & \textbf{80.24} & 90.84 & 76.44 & 90.00 \\
            {\hskip 0.5cm} + \textit{Test-Time Aug.} & \textit{81.52} & \textit{92.52} & \textit{78.36} & \textit{91.80} \\
            \textbf{Ours (Modified Model)} & \textbf{80.88} & \textbf{92.76} & 75.96 & \textbf{92.20} \\
            {\hskip 0.5cm} + \textit{Test-Time Aug.} & \textit{83.16} & \textit{94.28} & \textit{79.64} & \textit{94.20} \\
            \bottomrule
        \end{tabular}
    }
    \end{center}
    \caption{Test accuracy (\%) for Webvision. Top methods within 0.5\% in \textbf{bold}}
    \label{tab:webvision}
\end{table}

In Tables \ref{tab:animal10n_2} and \ref{tab:webvision}, we show the accuracy of our model on the Animal10N and Webvision benchmarks, respectively. In both cases, it can be seen that the modified model improves results, demonstrating that on these instance-dependant noise datasets, learning the relationship between images, noisy labels and `true' labels has allowed more accurate relabelling, and an improvement in final model accuracy. Animal10N, which consists of five pairs of commonly confused image classes, sees a significant improvement of $2.64\%$ from using our modified training approach, due to the relationship between commonly confused classes being easy to learn. Webvision sees a more modest but still meaningful improvement of $0.64\%$ accuracy when using our modified model.
It can be seen that on both benchmarks, we report state-of-the-art results, and that using dropout-based test-time augmentation further improves our accuracy.

We next turn our attention to the synthetic \textit{`Polynomial Margin Diminishing (PMD)'}~\cite{zhang2020learning} and \textit{`RoG'}~\cite{lee2019robust} instance-dependent noisy label benchmarks based on CIFAR10 and CIFAR100 in Tables \ref{tab:results_pmd_2} and \ref{tab:results_rog_2}. In all cases considered in these two tables, our accuracy results are substantially higher than by other approaches, even without our modified model. When we do use a modified model for these synthetic instance-dependent noisy-label benchmarks, we find mixed results, with performance often decreasing, potentially due to learned noise transitions not generalizing as well as they do in real-world instance-dependent noise datasets.

\begin{table*}[ht!]
    \begin{center}
    \scalebox{0.8}{
        \begin{tabular}{cc|c|c||c|c|c}
            \toprule
            \multicolumn{1}{c}{Dataset} & \multicolumn{3}{c}{CIFAR-10} & \multicolumn{3}{c}{CIFAR-100}\\    
            \midrule
            \multicolumn{1}{c}{Noise Type} & \multicolumn{1}{c}{Type-I 35\%} & \multicolumn{1}{c}{Type-II 35\%} & \multicolumn{1}{c}{Type-III 35\%} & \multicolumn{1}{c}{Type-I 35\%} & \multicolumn{1}{c}{Type-II - 35\%} & \multicolumn{1}{c}{Type-III - 35\%} \\
            \midrule
            Cross-Entropy & 78.11 & 76.65 & 76.89 & 57.68 & 57.83 & 56.07 \\
            PLC \cite{zhang2021learning} & 82.80 & 81.54 & 81.50 & 60.01 & 63.68 & 63.68 \\
            \midrule
            \textbf{Ours (Normal Model)} & \textbf{94.06} & \textbf{93.25} & \textbf{93.35} & 65.87 & 65.80 & 66.36 \\
            {\hskip 0.5cm} + \textit{Test-Time Aug.} & \textit{94.72} & \textit{93.79} & \textit{93.97} & \textit{66.83} & \textit{66.48} & \textit{67.42} \\
            \textbf{Ours (Modified Model)} & \textbf{94.00} & \textbf{93.76} & \textbf{94.23} & \textbf{68.25} & \textbf{68.14} & \textbf{68.22} \\
            {\hskip 0.5cm} + \textit{Test-Time Aug.} & \textit{94.39} & \textit{94.19} & \textit{94.23} & \textit{70.13} & \textit{69.35} & \textit{70.13} \\
            \bottomrule
        \end{tabular}
    }
    \end{center}
    \caption{Test accuracy (\%) for Polynomial Margin Diminishing Noise \cite{zhang2020learning}. Top methods are in \textbf{bold}.} 
    \label{tab:results_pmd_2}
\end{table*}

\begin{table*}[t!]
    \begin{center}
    \scalebox{0.8}{
        \begin{tabular}{ccccccc}
            \toprule
            Data set & \multicolumn{3}{c}{CIFAR-10} & \multicolumn{3}{c}{CIFAR-100}\\    
            \midrule
            Method/ Noise Ratio & DenseNet (32\%) & ResNet (38\%) & VGG (34\%) & DenseNet (34\%) & ResNet (37\%) & VGG (37\%) \\
            \midrule
            D2L + RoG~\citep{lee2019robust} &  68.57 & 60.25 & 59.94 & 31.67 & 39.92 & 45.42\\
            CE + RoG~\citep{lee2019robust} &  68.33 & 64.15 & 70.04 & 61.14 & 53.09 & 53.64\\
            PropMix~\citep{cordeiro2021propmix}& 84.25 & 82.51 & 85.74 & 60.98 & 58.44 & 60.01 \\
            \midrule
            \textbf{Ours (Normal Model)} & \textbf{93.26} & \textbf{92.05} & \textbf{93.29} & 62.47 & \textbf{64.91} & \textbf{64.98} \\
            {\hskip 0.5cm} + \textit{Test-Time Aug.} & \textit{93.87} & \textit{92.66} & \textit{93.86} & \textit{63.40} & \textit{65.74} & \textit{66.10} \\
            \textbf{Ours (Modified Model)} & 89.46 & 90.97 & 89.77 & \textbf{63.68} & 63.09 & 63.89 \\
            {\hskip 0.5cm} + \textit{Test-Time Aug.} & 90.25 & \textit{91.85} & \textit{90.42} & \textit{65.15} & \textit{64.70} & \textit{65.30} \\
            \bottomrule
        \end{tabular}
    }
    \end{center}
    \caption{Test accuracy (\%) for the RoG label noise benchmark~\citep{lee2019robust}, where baseline results are from~\citep{lee2019robust}. Top methods are in \textbf{bold}.} 
    \label{tab:results_rog_2}
\end{table*}

Finally, in Table \ref{tab:results_cifar_2} we show the results of our method on the synthetically constructed symmetric and asymmetric noise for CIFAR10 and CIFAR100. These noise types are rare in practice, but they are common noisy-label benchmarks so we include them here for completeness. We see that our method is competitive with the state-of-the-art on CIFAR10 symmetric and asymmetric noise, despite featuring fewer mechanisms designed to address these types of noise. We particularly note 40\% asymmetric noise, which benefits from the modified model due to noisy labels greatly limiting the set of feasible samples for each image, allowing us to exceed the state-of-the-art. In contrast to this, we report our results on symmetric label noise in CIFAR100, where existing methods perform better than ours. In all cases, our modified model is able to take advantage of the noise to provide more accurate relabelling, but the regularisation strategies that other methods use work well under the assumption of symmetric noise and provide stronger results. Perfectly symmetric noise over 100 classes is rare in practice though, and our results show universally strong performance on real-world instance-dependent datasets.

\begin{table*}[t!]
    \begin{center}
    \scalebox{0.8}{
        \begin{tabular}{l|cccc|c||cccc}
            \toprule
            \multicolumn{1}{c}{Dataset} & \multicolumn{5}{c}{CIFAR-10} & \multicolumn{4}{c}{CIFAR-100}\\    
            \midrule
            \multicolumn{1}{c}{Noise type} & \multicolumn{4}{c}{Sym.} & \multicolumn{1}{c}{Asym.} &  \multicolumn{4}{c}{Sym.} \\
            \midrule
            Method / Noise Ratio & {\hskip 0.2cm} 20\% {\hskip 0.1cm} & {\hskip 0.1cm} 50\% {\hskip 0.1cm} & {\hskip 0.1cm} 80\% {\hskip 0.1cm} & {\hskip 0.1cm} 90\% {\hskip 0.2cm} & {\hskip 0.2cm} 40\% {\hskip 0.2cm} & {\hskip 0.2cm} 20\% {\hskip 0.1cm} & {\hskip 0.1cm} 50\% {\hskip 0.1cm} & {\hskip 0.1cm} 80\% {\hskip 0.1cm} & {\hskip 0.1cm} 90\% {\hskip 0.2cm} \\
            \midrule
            Cross-Entropy \citep{li2020dividemix} & 82.7 & 57.9 & 26.1 & 16.8 & 72.3 &  61.8 & 37.3 & 8.8 & 3.5\\
            ELR~\citep{liu2020early} & \textbf{95.8} & \textbf{94.8} & 93.3 & 78.7 & 93.0 & 77.6 & 73.6 & 60.8 & 33.4 \\
            DivideMix \citep{li2020dividemix} & \textbf{95.7} & 94.4 & 92.9 & 75.4& 92.1  & 76.9 & 74.2 & 59.6 & 31.0 \\
            AugDesc \cite{Nishi_2021_CVPR} & \textbf{96.3} & \textbf{95.4} & \textbf{93.8} & 91.9 & 94.6 & \textbf{79.5} & \textbf{77.2} & 66.4 & 41.2 \\
            ContrastToDivide \cite{zheltonozhskii2022contrast} & \textbf{96.4} & \textbf{95.3} & \textbf{94.4} & \textbf{93.6} & 93.5 & \textbf{78.7} & \textbf{76.4} & \textbf{67.8} & \textbf{58.7} \\
            PropMix~\citep{cordeiro2021propmix} & \textbf{96.09} & \textbf{95.53} & \textbf{93.77} & \textbf{93.20} & 94.64 & 76.99 & 73.71 & 66.75 & \textbf{58.32}\\
            \midrule
            \textbf{Ours (Normal Model)} & 95.04 & \textbf{95.13} & \textbf{94.51} & 91.72 & 94.91 & 69.02 & 68.87 & 64.09 & 55.91 \\
            {\hskip 0.5cm} + \textit{Test-Time Aug.} & \textit{95.47} & \textit{95.39} & \textit{94.90} & \textit{92.32} & \textit{95.14} & \textit{70.17} & \textit{69.57} & \textit{65.10} & \textit{56.81} \\
            \textbf{Ours (Modified Model)} {\hskip 0.1cm} & \textbf{95.99} & \textbf{95.59} & \textbf{94.48} & \textbf{93.52} & \textbf{95.85} & 75.09 & 70.86 & 57.03 & 39.95 \\
            {\hskip 0.5cm} + \textit{Test-Time Aug.} & \textit{96.75} & \textit{96.16} & \textit{94.98} & \textit{93.93} & \textit{96.42} & \textit{76.65} & \textit{72.95} & \textit{57.83} & \textit{40.31} \\
            \bottomrule
        \end{tabular}
    }
    \end{center}
    \caption{Test accuracy (\%) for all competing methods on CIFAR-10 and CIFAR-100 under symmetric and asymmetric noises. Results from related approaches are as presented in \citep{li2020dividemix} and \cite{wu2021ngc}. Top methods within $1\%$ are in \textbf{bold}.} 
    \label{tab:results_cifar_2}
\end{table*}

\subsection{Predictions with Noisy Labels}

A unique feature of our method is that our final model can be used to predict the labels of samples with \textit{and} without noisy labels. In some applications, e.g., tagged image classification, images at test-time may also have noisy labels associated with them, which our model can use to improve classification performance. To show this, we generate artificial noisy labels for all the samples in the CIFAR10 test set using the same procedure as we used for 40\% Asymmetric noise, and in Table \ref{tab:label_ablation_accuracy} we show the accuracy of our model on these samples when these noisy labels are and are not used. We see that our model performance  increases when using the noisy labels, because our model has learned to use noisy labels (when they are available) to improve prediction accuracy.

\begin{table}[t!]
    \begin{center}
    \scalebox{0.9}{
        \begin{tabular}{lc|c}
            \toprule
            Test Set & {\hskip 0.02cm} & Accuracy \\
            \midrule
            Without Noisy Labels & & 95.85 \\
            With Noisy Labels & & 97.59 \\
            \bottomrule
        \end{tabular}
    }
    \end{center}
    \caption{\small Accuracy of our model tested with and without noisy labels on CIFAR10 Asym. 40\% noise. Noisy labels are generated for training and testing samples using the same procedure.} 
    \label{tab:label_ablation_accuracy}
\end{table}

We further demonstrate this learned relationship in Table \ref{tab:label_ablation_example}, where we show how the prediction of our model changes depending on what noisy label is provided to the model. In this 40\% Asymmetric noise dataset, the image of the dog shown would only ever be labelled as a dog or a cat, and in both cases it can be seen that our model makes the correct prediction. Our model also makes the correct prediction when provided with no noisy label, although with lower confidence because it does not have access to the noisy label. Because our model has learned that none of the other classes are ever confused with dogs, it makes other predictions when provided with these labels. These results show that our model has learned the noise relationship, and that it can use both the image features and the noisy label (when it is available) to generate higher confidence predictions.

\begin{table}[t!]
    \begin{center}
    \begin{minipage}{.25\columnwidth}
        \centering
        \includegraphics[width = 2.0cm]{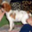}
    \end{minipage}
    \begin{minipage}{.6\columnwidth}
        \centering
        \scalebox{0.7}{
            \renewcommand{\arraystretch}{1.25}
            \begin{tabular}{ c|c } 
                \toprule
                Noisy Label & Prediction (Confidence) \\ 
                \midrule
                    \cellcolor{LightGrey} - & \cellcolor{LightGreen} Dog (90.27\%) \\
                    Airplane   & Bird       (65.39\%) \\
                    Automobile & Automobile (62.75\%) \\
                    Bird       & Bird       (98.30\%) \\
                    \cellcolor{LightGrey} Cat & \cellcolor{LightGreen} Dog (94.51\%) \\
                    Deer       & Deer       (96.06\%) \\
                    \cellcolor{LightGrey} Dog & \cellcolor{LightGreen} Dog (96.29\%) \\
                    Frog       & Frog       (96.49\%) \\
                    Horse      & Horse      (75.84\%) \\
                    Ship       & Ship       (95.46\%) \\
                    Truck      & Truck      (96.88\%) \\
                \bottomrule
            \end{tabular}
        }
    \end{minipage}
    \end{center}
    \caption{Predictions made by our model with different noisy labels for a testing sample (showing a dog) in Asym 0.4 noise for CIFAR10. `-' represents using a null label in place of a noisy label.}
    \label{tab:label_ablation_example}
\end{table}

\subsection{Clean Set Selection}

\begin{figure}[ht!]
    \begin{center}
    \includegraphics[width=4.1cm]{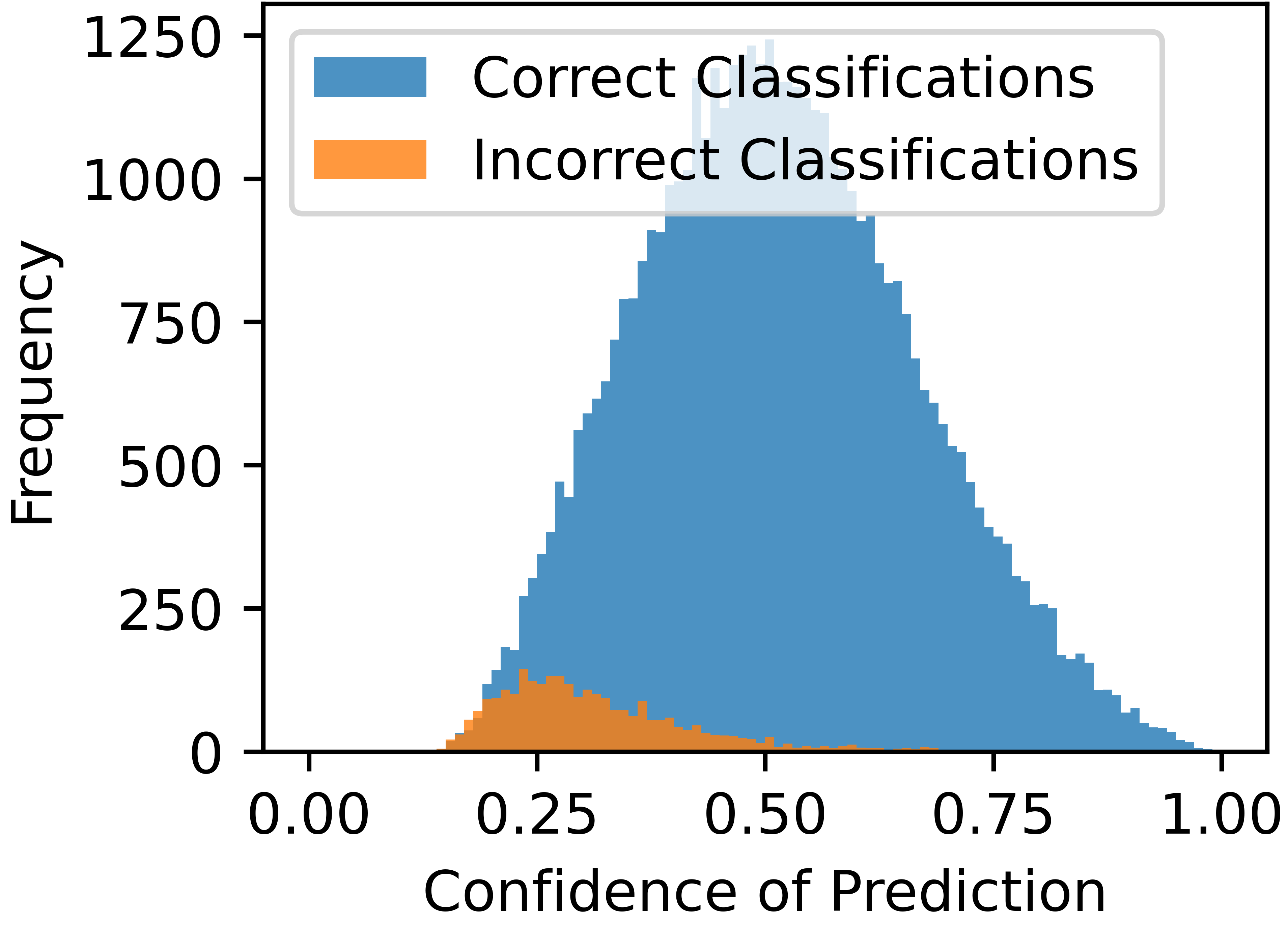}\hfill
    \includegraphics[width=3.9cm]{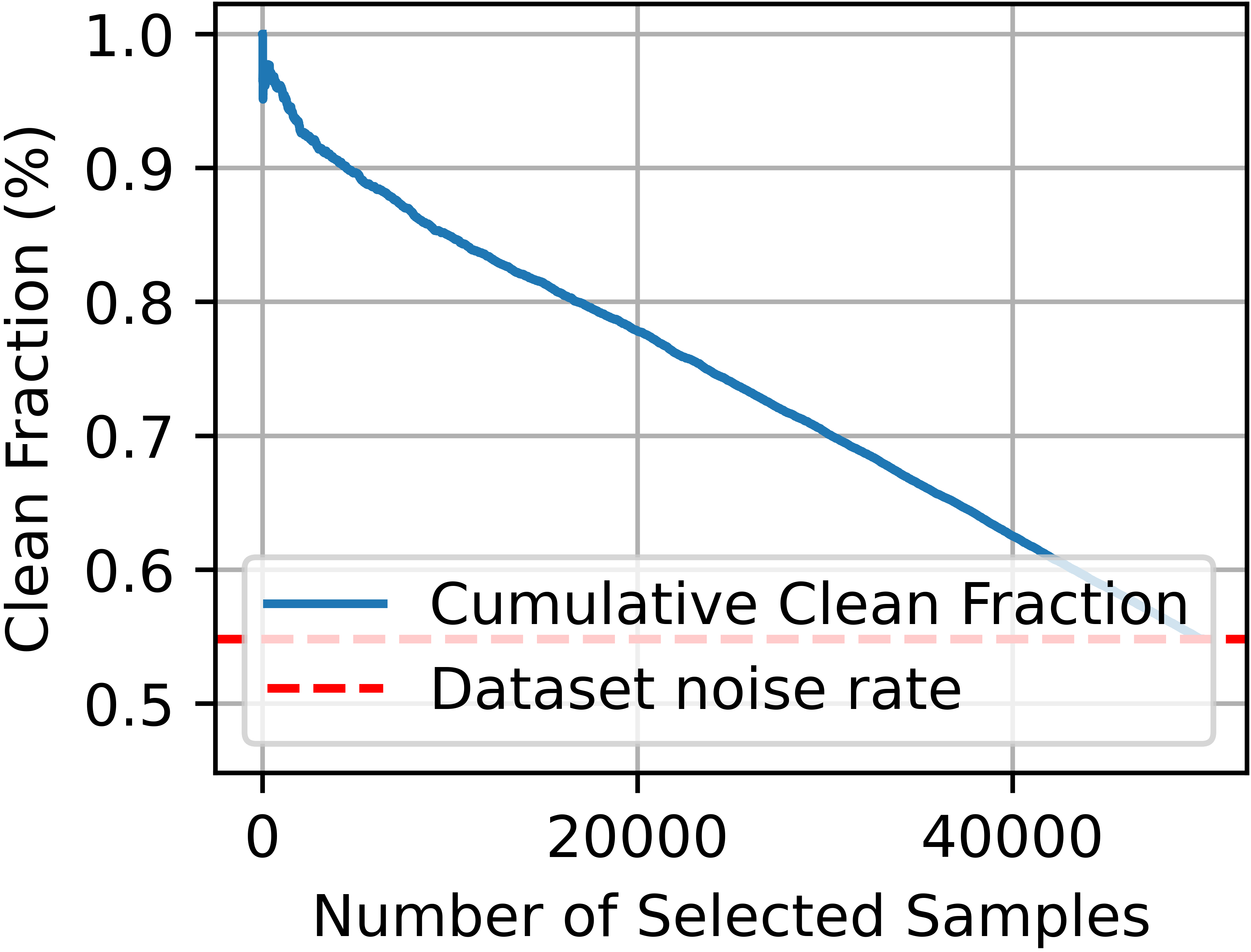}
    \end{center}
    \caption{Histograms showing the distribution of confidences after bootstrapping for correct/incorrect classifications (left), and how the highest confidence samples are disproportionally clean (right) for 50\% Symmetric Noise on CIFAR10.}
    \label{fig:cleanandcorrectbias}
\end{figure}

On the left of Figure \ref{fig:cleanandcorrectbias}, we show the distribution of confidences of our model after bootstrapping, and see that the highest confidence predictions are almost entirely for samples whose predicted `true' label ($\hat{\mathbf{y}}$) is correct, which allows us to select highly accurate clean sets for SSL. On the right, we show the percentage of the selected samples for the clean set that were originally clean (i.e. samples for which $\tilde{\mathbf{y}}$ matches the true label) if no form of noise balancing is performed. We see that if we did not perform noise balancing, the selected clean set would disproportionally consist of clean samples, which would cause a degenerate relationship between noisy labels and clean labels to be learned during the SSL stage of training, preventing accurate relabeling of samples from other noise transitions. 

\subsection{Ablations and Training Time}

In Tables \ref{tab:stage_accuracy}, \ref{tab:augmentation_ablation} and \ref{tab:self_supervision_ablation}, we perform a number of ablation studies on the CIFAR10 40\% Asymmetric noise. 
In Table \ref{tab:stage_accuracy}, we show the accuracy of our trained model after each stage of training. We see that the model accuracy improves after each stage of training, with the final training's use of MixUp and strong augmentations providing an additional $0.87\%$ accuracy over the semi-supervised learning stage.

In Table \ref{tab:augmentation_ablation}, we see the large impact that the choice of augmentations has on the number of errors in the clean set after the bootstrapping phase. We see that using strong augmentations for training greatly reduces the number of errors, likely due to their regularising effect and its ability to prevent overfitting to the noisy labels. During the evaluation stage however, averaging the prediction of the model over multiple weak augmentations performs best. We see that this matches the findings by Nishi et al.~\cite{Nishi_2021_CVPR}, who find that using strong augmentations for training and weaker augmentations for loss modelling works best.

In Table \ref{tab:self_supervision_ablation}, we see the similarly large impact that using self-supervised pretraining has on the number of errors in the clean set after bootstrapping. In the supplementary material, we show additional experiments with `null labels' and different model architectures.

As for training time on CIFAR10 problems, our method takes on average 13.8h for SimCLR pretraining, 0.5h for bootstrapping, 7.5h for SSL, and 2.5h for final training (total of 24.3h) on an Nvidia RTX 2080. In comparison, DivideMix~\cite{li2020dividemix} takes on average 5h, and the more recent method of PropMix~\cite{cordeiro2021propmix} can take up to 10h.

\begin{table}[t!]
    \begin{center}
    \scalebox{0.8}{
        \begin{tabular}{lc|c}
            \toprule
            Training Stage & {\hskip 0.02cm} & Accuracy \\
            \midrule
            After Bootstrapping & & 91.41 \\
            After Semi-Supervised Learning & & 94.98 \\
            After Final Training & & 95.85 \\
            \bottomrule
        \end{tabular}
    }
    \end{center}
    \caption{Model accuracy after each stage of training on CIFAR10 Asym. 40\% noise.}
    \label{tab:stage_accuracy}
\end{table}

\begin{table}[t!]
    \begin{center}
    \scalebox{0.8}{
        \begin{tabular}{clc|ccc}
            \toprule
             & &  {\hskip 0.1cm} & \multicolumn{3}{c}{Evaluation Aug.} \\
             & &  {\hskip 0.1cm} & {\hskip 0.2cm} None {\hskip 0.1cm} & {\hskip 0.1cm} Weak {\hskip 0.1cm} & {\hskip 0.1cm} Strong {\hskip 0.2cm} \\
            \midrule
            \parbox[t]{2mm}{\multirow{3}{*}{\rotatebox[origin=c]{90}{Train}}} {\hskip 0.1cm} & None & {\hskip 0.1cm} & 579 & 361 & 456 \\
            & Weak & {\hskip 0.1cm} & 265 & 56 & 300 \\
            & Strong & {\hskip 0.1cm} & 28 & \textbf{21} & 31 \\
            \bottomrule
        \end{tabular}
    }
    \end{center}    
    \caption{Effect of different training/testing augmentations on the number of errors in a clean set of 10,000 samples selected after bootstrapping. Test performed on CIFAR10 Asym. 40\% noise.}
    \label{tab:augmentation_ablation}
\end{table}

\begin{table}[t!]
    \begin{center}
    \scalebox{0.8}{
        \begin{tabular}{lc|c}
            \toprule
            Training Strategy & {\hskip 0.1cm} & {\hskip 0.1cm} No. of Errors \\
            \midrule
            No Self-Supervision & {\hskip 0.1cm} & 369 \\
            With Self-Supervision & {\hskip 0.1cm} & 21 \\
            \bottomrule
        \end{tabular}
    }
    \end{center}
    \caption{Effect of self-supervision on the number of errors in a clean set of 10,000 samples selected after bootstrapping. Test performed on CIFAR10 Asym. 40\% noise}
    \label{tab:self_supervision_ablation}
\end{table}

\section{Conclusions}

In this paper, we proposed a new method that predicts labels from images and their noisy labels. Unlike other methods, our training procedure does not require access to a clean set of data, which we achieve by introducing bootstrapping and a careful noise-based balancing procedure. By utilising the relationship between images, noisy labels and `clean' labels to accurately relabel samples, we find that we can achieve SOTA results. By simply changing the model used, we further unify the noisy label learning and semi-supervised learning domains, resulting in a simplified architecture that can improve performance for challenging instance-dependent noisy-label tasks. Additionally, we find that by randomly replacing noisy labels with `null' labels during training, we can remove the need for model distillation, allowing practitioners of our method to perform predictions with and without noisy labels. 

\clearpage

{\small
\bibliographystyle{ieee_fullname}
\bibliography{egbib}
}

\clearpage

\appendix

\section{Additional Training Information}

\subsection{Self-Supervised Pretraining}

For CIFAR-10, CIFAR100 and Animal10N we use SimCLR~\cite{chen2020simple} for self-supervised pre-training. Pretraining is done for 1000 epochs, with the learning rate starting at 0.5, and decaying by a factor of 0.1 after 700, 800 and 900 epochs. We use stochastic gradient descent as our optimizer, with Nesterov momentum of 0.9 and weight decay of $1 \times 10^{-4}$. We use a temperature of 0.5 for SimCLR and a batch size of 512.

For Webvision, we adopt MoCo-v2~\cite{chen2020improved}, trained for 100 epochs (with 1 epoch of warmup) and with a batch size of 128. We use stochastic gradient descent as our optimizer, with momentum of 0.9 and weight decay of $1 \times 10^{-4}$. The learning rate starts at 0.015, decaying by a factor of 0.1 at epoch 50.

The feature embeddings generated by our models have 512 dimensions.

\subsection{Bootstrapping Training}

For CIFAR10, CIFAR100 and Animal10N, we do 60 epochs of bootstrapping training with MixUp. We use a learning rate of 0.02, which decays to 0.002 after 5 epochs and to 0.0002 after 50 epochs. Stochastic gradient descent is used as the optimizer, with Nesterov momentum of 0.9 and weight decay of $5 \times 10^{-4}$.

For Webvision, we do 300 epochs of bootstrapping training with MixUp. We use a starting learning rate of 0.005, which increases linearly for the first 30 epochs until it reaches 0.1, and then follows a cosine learning rate decay (capped at a minimum of $1 \times 10^{-5}$). Stochastic gradient descent is used as the optimizer, with Nesterov momentum of 0.9 and weight decay of $1 \times 10^{-5}$.

In all cases, we use a batch size of 64 and a MixUp alpha of 0.2.

\subsection{Semi-Supervised Learning}

For semi-supervised learning, we use FixMatch for all our experiments, with the temperature set at 0.5, the confidence threshold for pseudo-label generation set at 0.95, and the unlabelled loss ratio set at 1.0. We train for 100,000 iterations in all cases, and use an exponential moving average momentum of 0.999.

We use a cosine learning rate, starting at 0.02 (capped at a minimum of $1 \times 10^{-5}$ for Webvision, and at $1 \times 10^{-4}$ for all other experiments). We use stochastic gradient descent with nesterov momentum of 0.9, and weight decay of $1 \times 10^{-5}$ for Webvision and $5 \times 10^{-4}$ for all other experiments.

For CIFAR10, CIFAR100 and Animal10N, we use a batch size of $64$ clean samples, and $3 \times 64$ noisy samples per batch. For Webvision, we use a batch size of $32$ clean samples, and $3 \times 32$ noisy samples per batch.

\subsection{Final Model Training}

For final model training, we do 300 epochs of training for all experiments.

We use a cosine learning rate, starting at 0.02 (capped at a minimum of $1 \times 10^{-5}$ for Webvision, and at $1 \times 10^{-4}$ for all other experiments). We use stochastic gradient descent with nesterov momentum of 0.9, and weight decay of $1 \times 10^{-5}$ for Webvision and $5 \times 10^{-4}$ for all other experiments.

We use a batch size of 64 for Webvision and 128 for all other experiments,

\subsection{Creating the Clean, Noisy and Final Datasets}

For all of our experiments, we generate predictions for samples by averaging over 25 weak augmentations of each sample, and use the 90\% most confident predictions to estimate the noise transition matrix for the dataset.

For CIFAR10 and Animal10N, we set $K=0.1$ and $\tau=0.99$. For CIFAR100 and Webvision, we set $K=0.25$ and $\tau=0.99$.

\subsection{Model Architecture}

In the modified networks that we used to learn the relationship between images, noisy labels and true labels, we project both the images and noisy labels to have an encoding size of 128 before concatenating them together, with our hidden layer also having a size of 128. We use a dropout layer with $p=0.2$ after each of these linear projections (except the final classifier head), and we use batch normalization before the final classifier head. 

\section{Effect of Null Label Type}

In our method, we describe the use of a `null' label to represent the case where no noisy label is present. For all of our experiments, if there are $k$ classes in the training set, we use a $k$-wide zero vector as our `null' label. Here, we experiment with two alternative choices:

\begin{itemize}
    \item One Vectors: Using a $k$-wide vector filled with ones;
    \item $\frac{1}{k}$ Vectors: Using a $k$-wide vector where every value equals $\frac{1}{k}$ (so that the sum of all values is 1).
\end{itemize}

We show the results of using these alternative `null' label representations in Table \ref{tab:null_label_ablation}.

\begin{table}[ht!]
    \begin{center}
        \scalebox{1.0}{
            \begin{tabular}{c|c}
                \toprule
                Null Label Type & Accuracy \\
                \midrule
                Zero Vectors & 95.70 \\
                One Vectors & 95.82 \\
                $\frac{1}{k}$ Vectors & 95.50 \\
                \bottomrule
            \end{tabular}
        }
    \end{center}
    \caption{\small Accuracy using different null label methods for Asym. 40\% noise on CIFAR10} 
    \label{tab:null_label_ablation}
\end{table}

In our experiments, we find that the choice of null label representation has little impact on the final performance of the model. In all cases, the model is able to learn to make predictions when a noisy label is and is not present.

\section{Ablation of Model Construction}

For all of our results, we use a `concatenation' based model architecture, where image features and noisy labels are combined by projecting them to the same dimensionality, and then concatenating them together before passing them through the remaining linear, ReLU and batch normalisation layers of the network. This form of combining noisy labels and image features together with concatenation is the standard method used by contemporary works~\cite{gu2021instancedependent,inoue2017multi,veit2017learning}.

Here, we briefly explore two other potential architectures for combining image features and noisy label information together.

\begin{itemize}
    \item Mixture of Experts: A separate classification head is created for every noisy label class, with the noisy label controlling which noisy label head is used for the prediction. In the case of mixed noisy labels (such as when performing MixUp between two noisy labels of different classes), the model output is the linear combination of each of the classification heads, weighted by their corresponding value in the noisy label;
    \item Attention: Scaled dot-product attention, as described by Vaswani et al.~\cite{vaswani2017attention}, is used to allow different noisy labels to attend to different image features. For our experiments on CIFAR10, we generate a query by projecting the noisy label to a $1 \times 16$ tensor, generate keys by projecting image features to a $128 \times 16$ tensor (representing a set of 128 keys), and generate values by projecting image features to a $128 \times 16$ tensor (representing a set of 128 values). Scaled dot-product attention is then used to compute a $1 \times 16$ feature tensor, with a final linear layer acting as a classification head.
\end{itemize}

In Table \ref{tab:model_ablation}, we show the results obtained by our training method using all three of these model architectures on 40\% Asymmetric noise on CIFAR10. We see that the concatenation and mixture of experts models perform similarly well, with the attention based model performing $\sim 1.5\%$ worse.

\begin{table}[ht!]
    \begin{center}
        \scalebox{1.0}{
            \begin{tabular}{c|c}
                \toprule
                Model Type & Accuracy \\
                \midrule
                Concatenation & 95.65 \\
                Mixture of Experts & 95.53 \\
                Attention & 94.02 \\
                \bottomrule
            \end{tabular}
        }
    \end{center}
    \caption{\small Accuracy using different model constructions for Asym. 40\% noise on CIFAR10} 
    \label{tab:model_ablation}
\end{table}

We note however that our exploration into using these model types is limited, and there may be opportunities to further optimise for these architectures.

\section{Effect of Dropping Labels for Pseudo-Label Generation}

In Section 3.2, we discuss how label-dropping is used during semi-supervised learning to allow our model to make predictions with and without noisy labels present. However, we do not use label dropping for the weakly augmented samples used for pseudo-label generation, with the justification that the model loss is not backpropagated through the weakly augmented samples in the FixMatch algorithm, and that always using noisy labels improves pseudo-label accuracy. Here, we experimentally justify this decision by comparing the final model accuracy when label dropping is and is not used for weakly augmented samples during semi-supervised learning.

\begin{table}[ht!]
    \begin{center}
        \scalebox{1.0}{
            \begin{tabular}{c|c}
                \toprule
                - & Accuracy \\
                \midrule
                With Label Dropping & 95.33 \\
                Without Label Dropping & 95.74 \\
                \bottomrule
            \end{tabular}
        }
    \end{center}
    \caption{\small Effect of Label Dropping on Final Accuracy for Asym. 40\% noise on CIFAR10} 
    \label{tab:label_dropping_ablation}
\end{table}

Here, we see that using label dropping for weakly augmented samples decreases the accuracy of the model. Thus, we always use noisy labels for pseudo-label generation.

\section{Generating Plausible Noisy Labels for Testing Samples}

In our experiments, we explore using `null' labels in place of a noisy label for samples at testing time. However, rather than passing a null label into the model alongside the testing sample, we could attempt to generate plausible `noisy' labels from testing samples.

In this experiment, we attempt to generate plausible noisy labels from samples using the model as it was at the end of the bootstrapping phase. For a given testing sample, we use the bootstrapping model to generate the `noisy' label, then we pass the testing sample and the `noisy' label into the final trained model to generate the final prediction.

\begin{table}[ht!]
    \begin{center}
        \scalebox{1.0}{
            \begin{tabular}{c|c}
                \toprule
                - & Accuracy \\
                \midrule
                With Null Labels & 95.74 \\
                With Label Generation & 95.09 \\
                \bottomrule
            \end{tabular}
        }
    \end{center}
    \caption{\small Comparison of Null Labels and Label Generation on Final Accuracy for Asym. 40\% noise on CIFAR10} 
    \label{tab:label_generation_ablation}
\end{table}

Here, we see that attempting to generate plausible noisy labels is a less effective strategy that using null labels to represent samples without associated noisy labels.

\section{70\% PMD Noise on CIFAR10 and CIFAR100}

In this section, we investigate the results of our method on 70\% PMD noise for CIFAR10 and CIFAR100. In Table \ref{tab:results_pmd_3}, we show the accuracy of our method on these datasets. We see that on CIFAR100 we get SOTA results, greatly surpassing the existing PLC method. However, on CIFAR10, we perform poorly.

\begin{figure}[!ht]
    \begin{center}
    \subfigure[PMD-1-0.35 \newline Noise Transition Matrix]{\includegraphics[width=4cm]{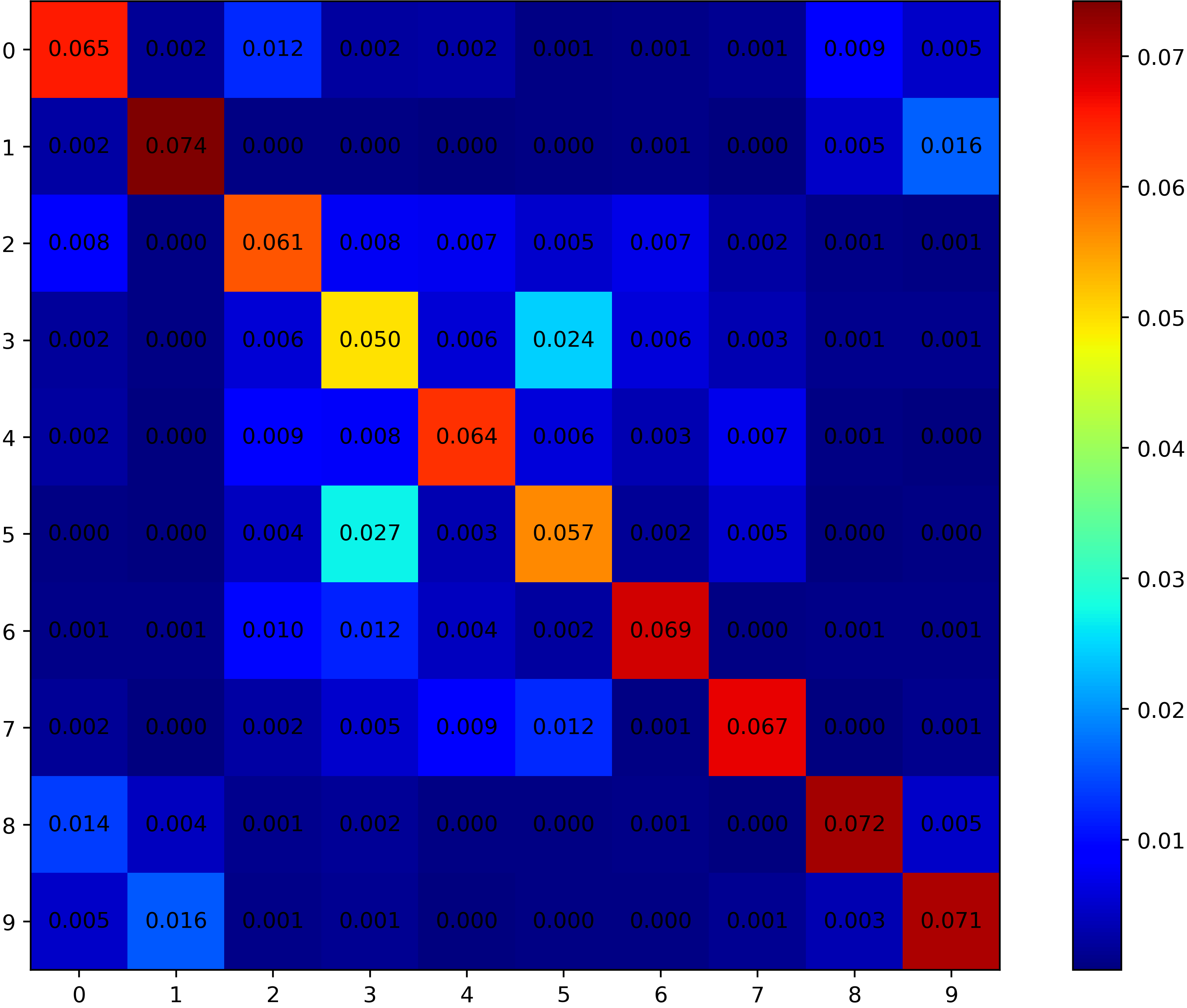}}
    \subfigure[PMD-1-0.35 \newline Final Confusion Matrix]{\includegraphics[width=4cm]{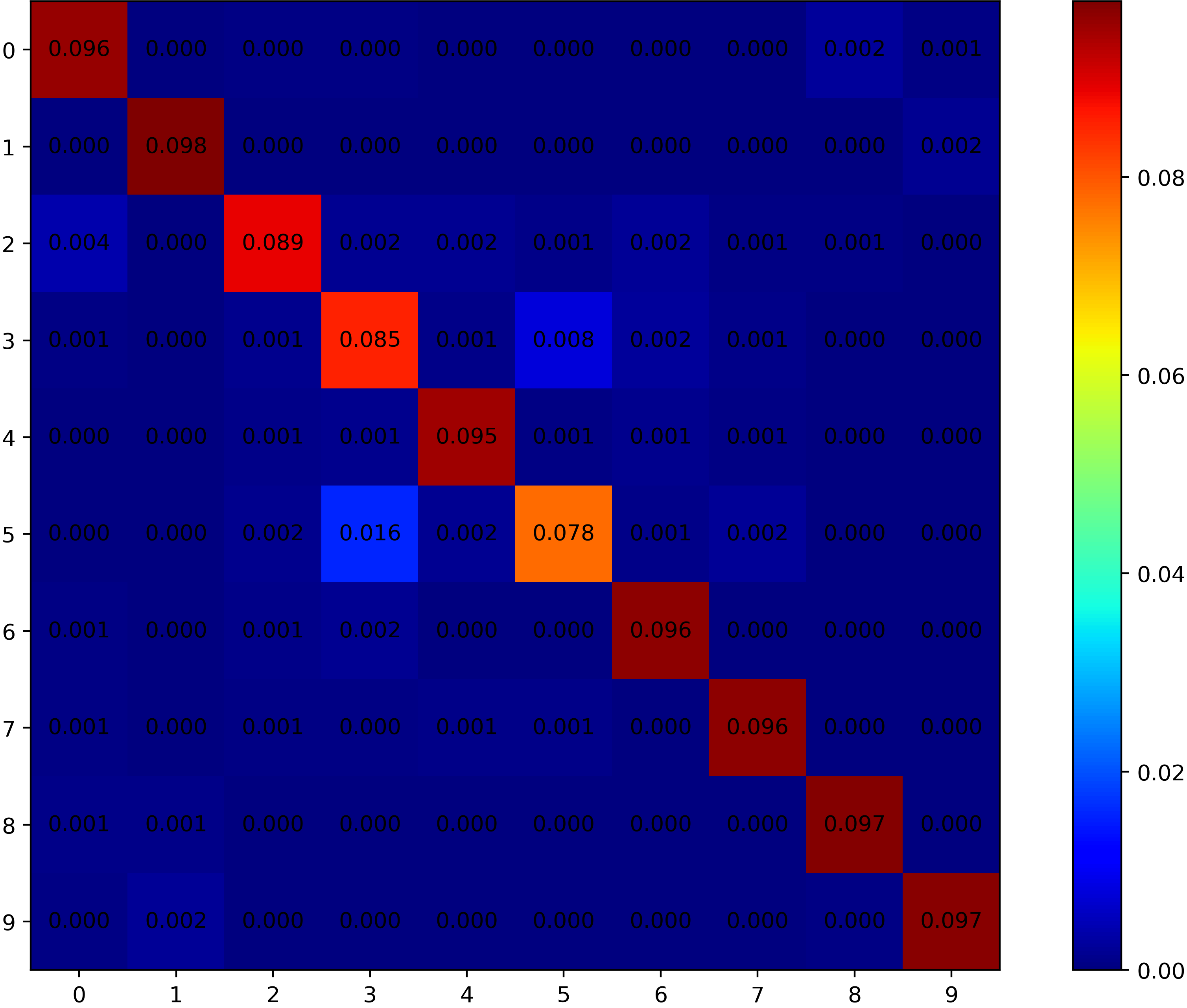}}
    \subfigure[PMD-1-0.70 \newline Noise Transition Matrix]{\includegraphics[width=4cm]{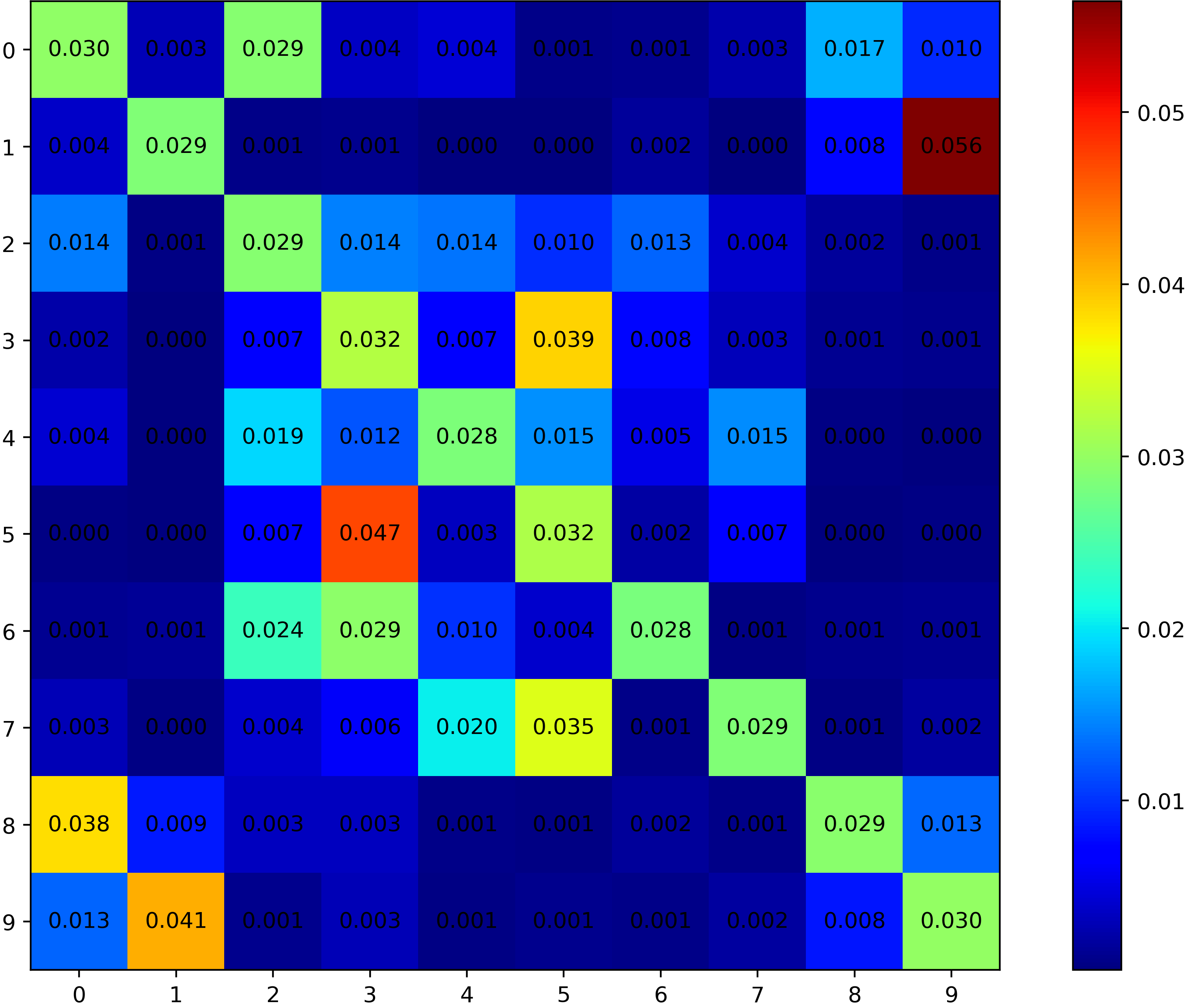}}
    \subfigure[PMD-1-0.70 \newline Final Confusion Matrix]{\includegraphics[width=4cm]{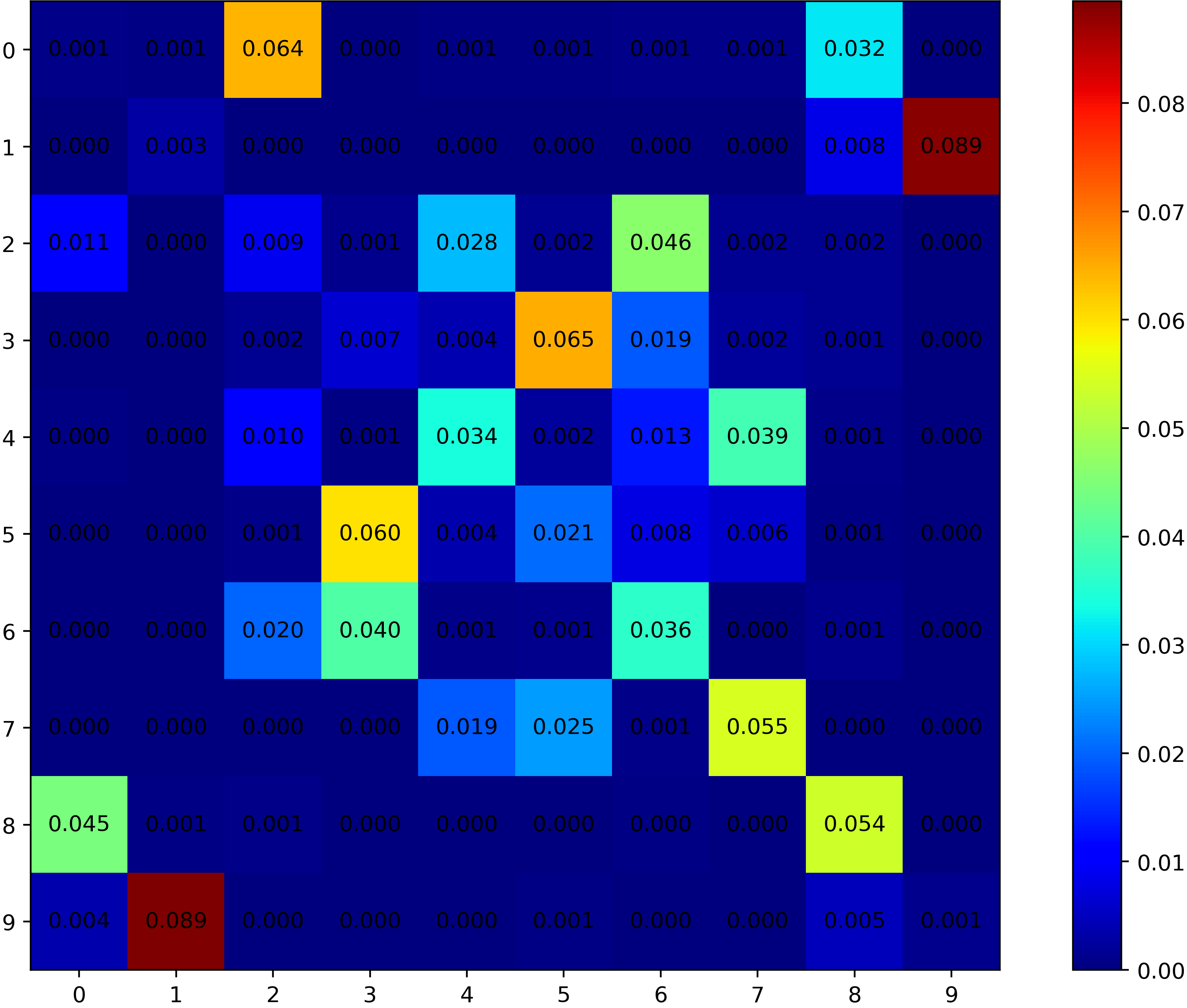}}
    \end{center}
    \caption{Noise Transition and Confusion Matrices for PMD-1 Noise on CIFAR10}
    \label{fig:cifar10-pmd}
\end{figure}

\begin{figure}[!ht]
    \begin{center}
    \subfigure[PMD-1-0.35 \newline Noise Transition Matrix]{\includegraphics[width=4cm]{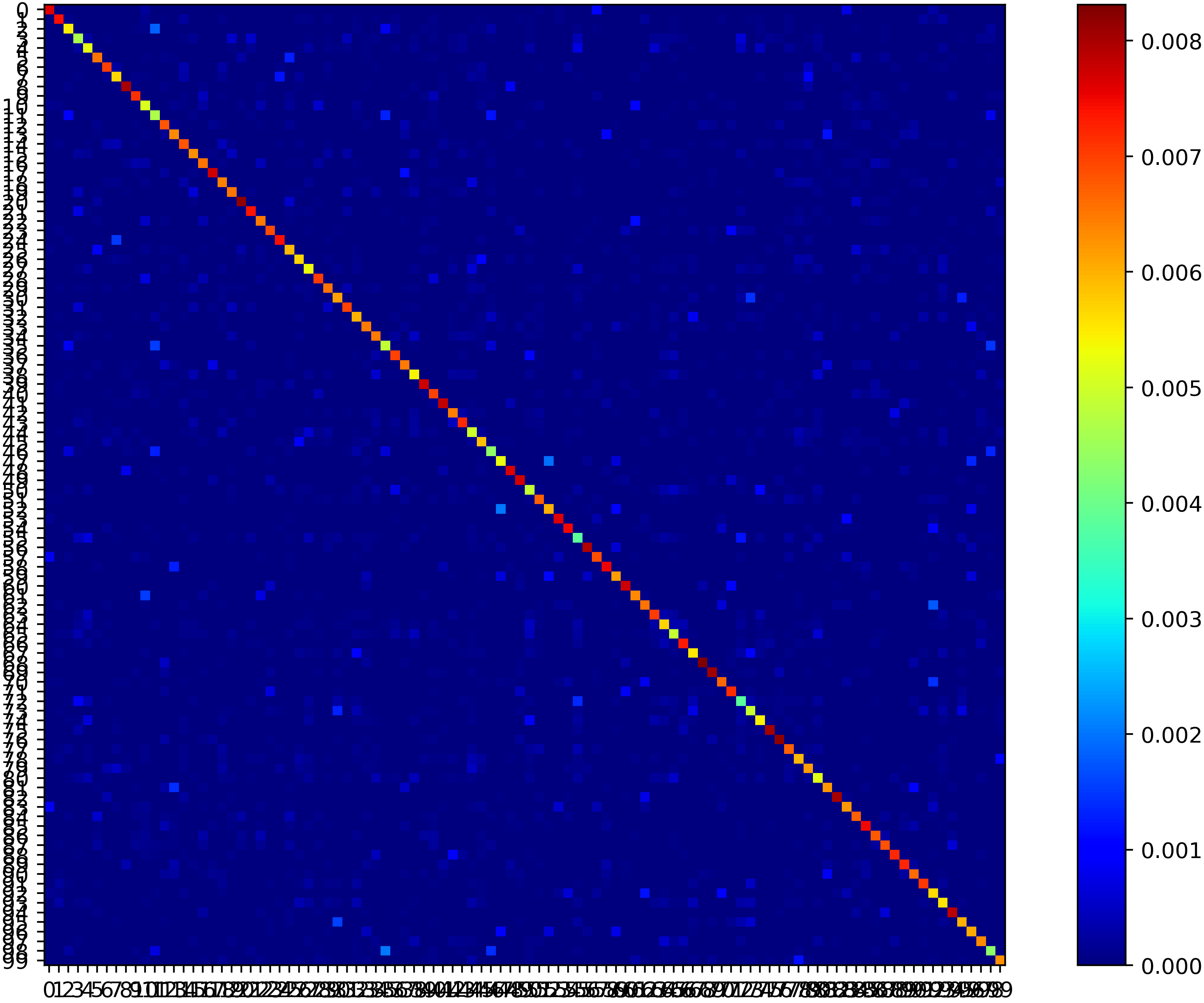}}
    \subfigure[PMD-1-0.35 \newline Final Confusion Matrix]{\includegraphics[width=4cm]{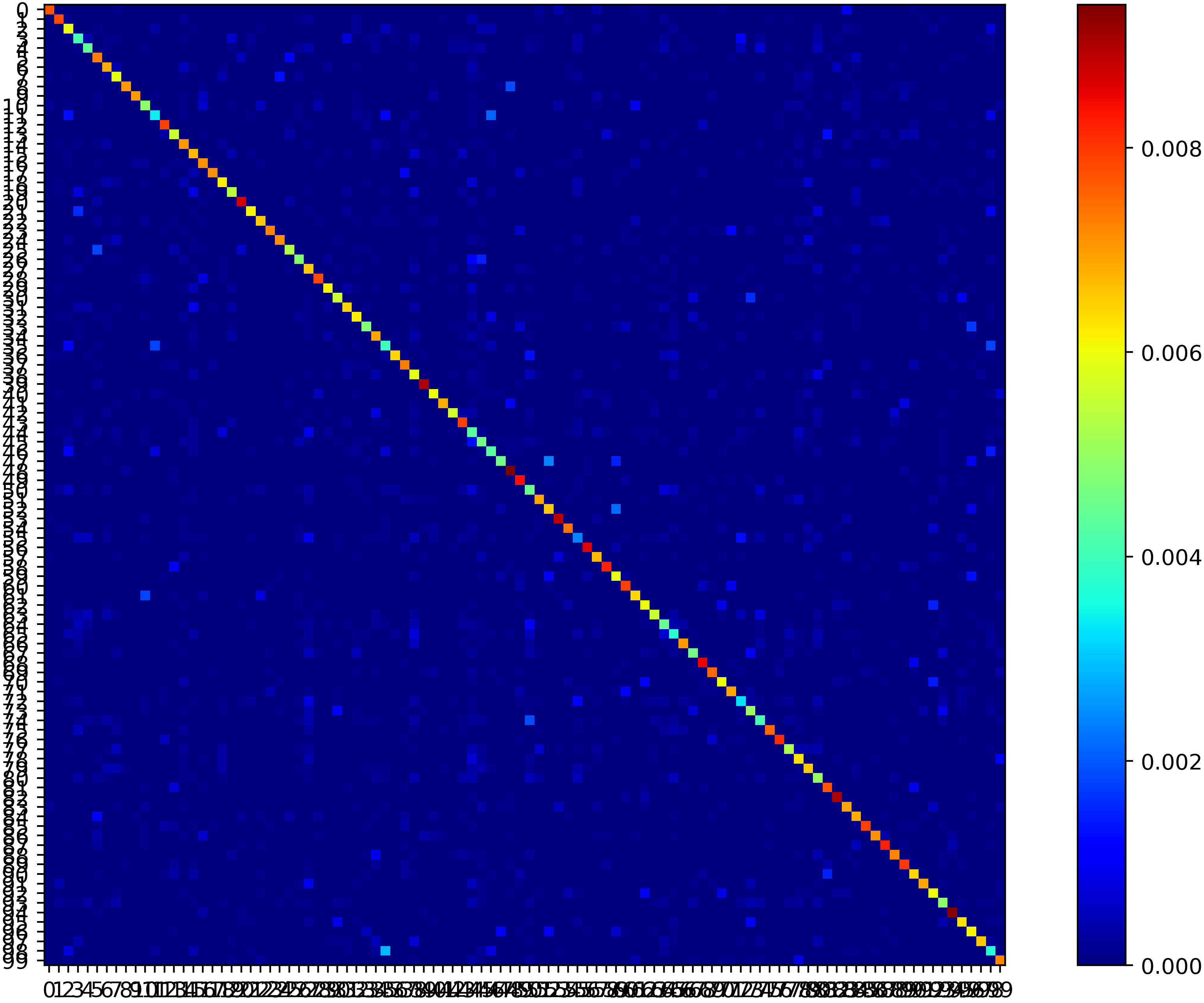}}
    \subfigure[PMD-1-0.70 \newline Noise Transition Matrix]{\includegraphics[width=4cm]{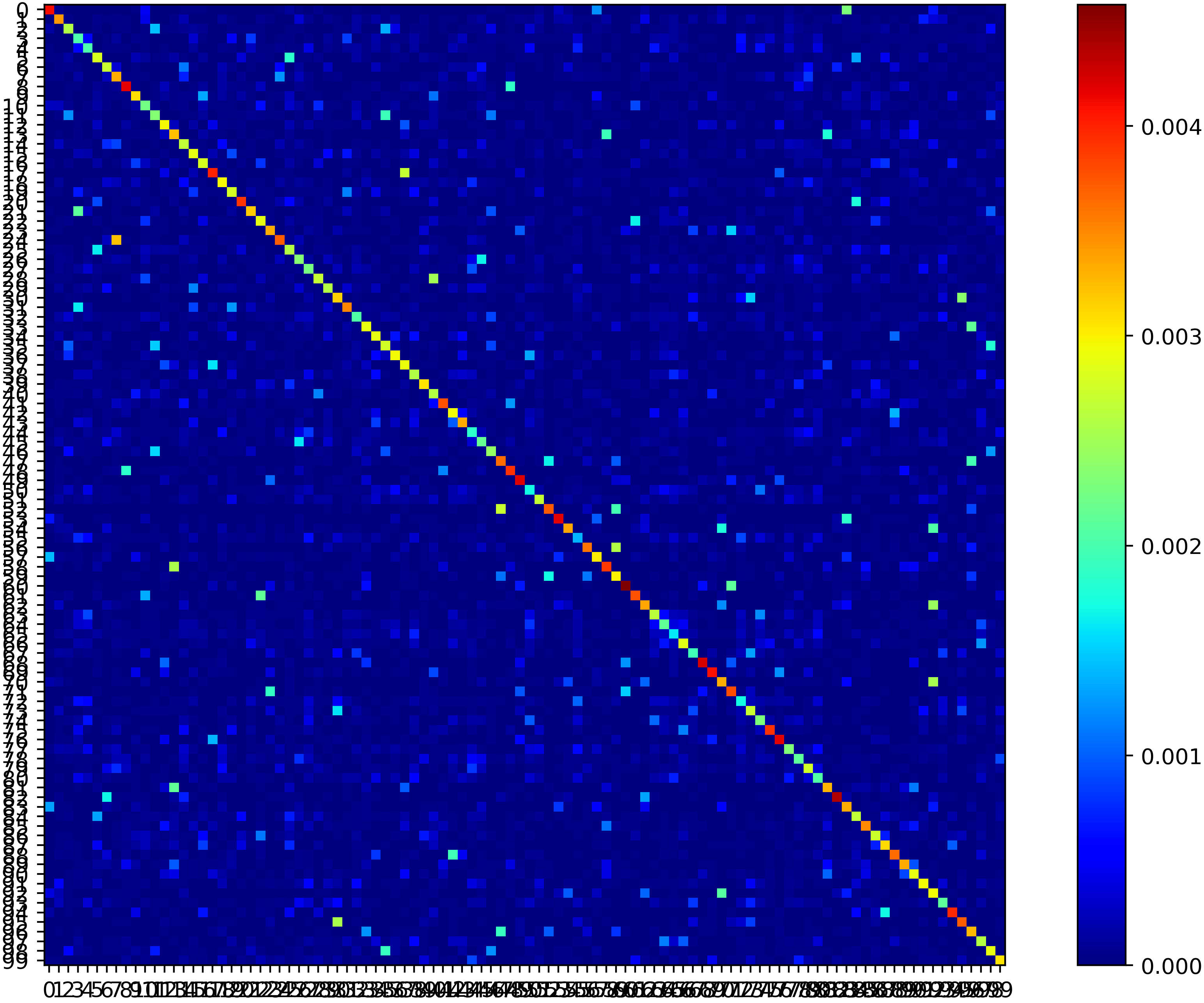}}
    \subfigure[PMD-1-0.70 \newline Final Confusion Matrix]{\includegraphics[width=4cm]{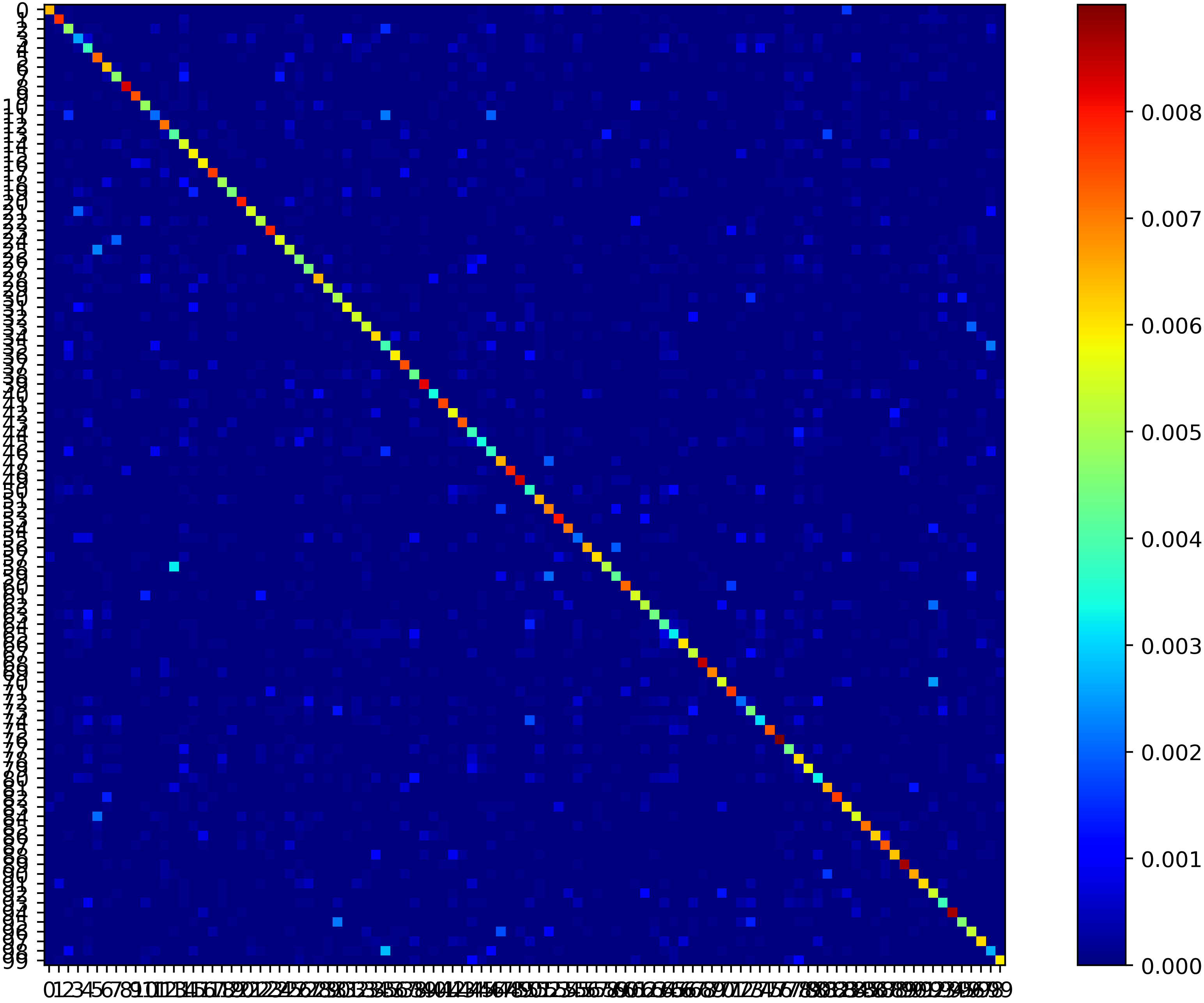}}
    \end{center}
    \caption{Noise Transition and Confusion Matrices for PMD-1 Noise on CIFAR100}
    \label{fig:cifar100-pmd}
\end{figure}

To understand this, in Figures \ref{fig:cifar10-pmd} and \ref{fig:cifar100-pmd} we show the noise transition matrix and the final confusion matrix of our model for 35\% and 70\% PMD-1 noise for CIFAR10 and CIFAR100.

In Figure \ref{fig:cifar10-pmd}(c), we see the noise transition matrix for PMD-1-0.70 Noise on CIFAR10, and we note that by introducing 70\% noise in a non-symmetric way, several of the classes have `flipped'. For example, more cats have become labelled as dogs than are labelled as cats, and vice versa (classes 3 and 5). Trucks and airplanes have similarly become flipped. Because of this, our method attempts to `correct' samples to the wrong class, which we show in Figure \ref{fig:cifar10-pmd}(d).

If we measure the performance of our model with respect to the flipped classes (by associating each label with the modal class it represents in the noise transition matrix), we find that our model has an accuracy of 45.7\%, slightly surpassing the accuracy of PLC.

On CIFAR100, this form of class flipping happens much more rarely due to the 70\% of mislabelled samples being `spread out' among over more classes. Because of this, our method is able to achieve much higher accuracy on CIFAR100 than it does in CIFAR10. In Figure \ref{fig:cifar100-pmd}(d), we show that the final confusion matrix for our trained model on CIFAR100 PMD-1-0.70 noise is much cleaner than it is for CIFAR10 PMD-1-0.70 noise (shown in Figure \ref{fig:cifar10-pmd}(d)).

\begin{table*}[ht!]
    \begin{center}
    \scalebox{0.9}{
        \begin{tabular}{cc|c|c||c|c|c}
            \toprule
            \multicolumn{1}{c}{Dataset} & \multicolumn{3}{c}{CIFAR-10} & \multicolumn{3}{c}{CIFAR-100}\\    
            \midrule
            \multicolumn{1}{c}{Noise Type} & \multicolumn{1}{c}{Type-I 70\%} & \multicolumn{1}{c}{Type-II 70\%} & \multicolumn{1}{c}{Type-III 70\%} & \multicolumn{1}{c}{Type-I 70\%} & \multicolumn{1}{c}{Type-II - 70\%} & \multicolumn{1}{c}{Type-III - 70\%} \\
            \midrule
            Cross-Entropy & 41.98 & 45.57 & 43.42 & 39.32 & 39.30 & 40.01 \\
            PLC \cite{zhang2021learning} & 42.74 & 46.04 & 45.05 & 45.92 & 45.03 & 44.52 \\
            \midrule
            \textbf{Ours (Regular Model)} & 21.02 & 27.55 & 21.27 & 58.15 & 53.77 & 57.81 \\
            {\hskip 0.5cm} + \textit{Test-Time Aug.} & \textit{21.17} & \textit{27.38} & \textit{20.99} & \textit{59.27} & \textit{54.44} & \textit{58.77} \\
            \textbf{Ours (Modified Model)} & 19.18 & 27.28 & 19.94 & \textbf{58.69} & \textbf{58.03} & \textbf{57.90} \\
            {\hskip 0.5cm} + \textit{Test-Time Aug.} & \textit{19.71} & \textit{27.17} & \textit{20.21} & \textit{59.39} & \textit{58.95} & \textit{58.95} \\
            \bottomrule
        \end{tabular}
    }
    \end{center}
    \caption{Test accuracy (\%) for Polynomial Margin Diminishing Noise \cite{zhang2020learning}. Top methods are in \textbf{bold}.} 
    \label{tab:results_pmd_3}
\end{table*}

\end{document}